%% file: paper.tex
\documentclass[]{moworld}
\microtypesetup{expansion=false}
\usepackage[T1]{fontenc}
\usepackage[toc,page,header]{appendix}
\usepackage{wrapfig}
\usepackage{multirow}
\usepackage{makecell}
\usepackage{xcolor}
\usepackage{colortbl}
\usepackage{amssymb}

\input{commands.tex}
\input{math_commands.tex}

\definecolor{colorfirst}{rgb}{.866,.945, 0.831}
\definecolor{colorsecond}{rgb}{1, 0.98, 0.83}
\definecolor{colorthird}{rgb}{0.76, 0.87, 0.92}
\definecolor{colorcite}{rgb}{0.212, 0.490, 0.741}

\usepackage{cleveref}
\crefname{figure}{Fig.}{Figs.}
\crefname{table}{Tab.}{Tabs.}
\crefname{equation}{Eq.}{Eqs.}
\crefname{section}{Sec.}{Secs.}

\title{MoWorld: A Flash World Model}

\author{MoWorld Team, Moxin Technology}
\abstract{\input{sections_en/0-abstract-english}}

\begin{document}

\maketitle

\input{sections_en/1-intro-english}
\input{sections_en/2-data-engine}
\input{sections_en/3-pretrain-english}
\input{sections_en/4-distillation-english-simple}
\input{sections_en/5-inference-english}

\input{sections_en/6-evaluation-english}

\input{sections_en/7-application-english}

\input{sections_en/8-conclusion-english}
\input{sections_en/9-contribution}

\clearpage
\bibliographystyle{unsrtnat}
\bibliography{main}
\clearpage



\end{document}

%% file: math_commands.tex
\usepackage{amsmath,amsfonts,bm}

\usepackage{xspace}

\newcommand{\markedblanklines}[2][\textbullet]{
  \par
  \begingroup
  \setlength{\parskip}{0pt}
  \setlength{\parindent}{0pt}
  \count0=0
  \loop
    \ifnum\count0<#2
      #1\hspace{1em}\rule{0pt}{\baselineskip}\par
      \advance\count0 by 1
  \repeat
  \endgroup
}

\def\eqref#1{equation~\ref{#1}}

\def\1{\bm{1}}

\DeclareMathAlphabet{\mathsfit}{\encodingdefault}{\sfdefault}{m}{sl}
\SetMathAlphabet{\mathsfit}{bold}{\encodingdefault}{\sfdefault}{bx}{n}

%% file: sections_en/0-abstract-english.tex
The future of World Models depends not only on scaling model capability, but also on scaling practicality and inference efficiency. High-frame-rate inference enables responsive perception, planning, and control in real-world autonomous systems. To this end, we present MoWorld, a cost-effective yet high-performance Flash World Model with an end-to-end framework spanning data generation, pre-training, distillation, and efficient inference, enabling up to\textbf{ 50 FPS real-time interaction with cinematic visual quality without the need of high-end GPUs}. To enable large-scale real-world deployment, MoWorld jointly optimizes model capability and cost throughout the entire development pipeline. Specifically, unlike existing approaches that primarily rely on large-scale video corpora, MoWorld is built upon a scalable 3D-native data engine accumulated from our large-scale 3D vision and generative modeling pipeline, enabling the efficient construction of geometrically consistent training data across diverse real-world and synthetic environments. Based on this foundation, a curriculum cross-frame pre-training strategy for stable and scalable World Model learning, an efficient denoising-step distillation algorithm to reduce diffusion training cost, and a mixed-precision parallel inference framework for low-cost real-time deployment. MoWorld is the first real-time interactive World Model built on the Neural Processing Unit (NPU) and can achieves up to 50 FPS in such the devices, enabling practical and efficient deployment at scale. Comprehensive evaluations demonstrate that MoWorld achieves leading performance; notably, its average inference cost is only 30\%-50\% of that of existing World Models, providing a practical foundation for large-scale real-world applications of World Models. We also demonstrate diverse applications of MoWorld, include Video Style Transfer, Video Editing, Point Cloud Reconstruction, Gaussian Splatting and more. 

~\\
Project Page: \url{https://moxin-tech.github.io/moworld/}

%% file: sections_en/1-intro-english.tex
\section{Introduction}

\begin{figure}[t]
\begin{center}
  \includegraphics[alt={},width=\textwidth]{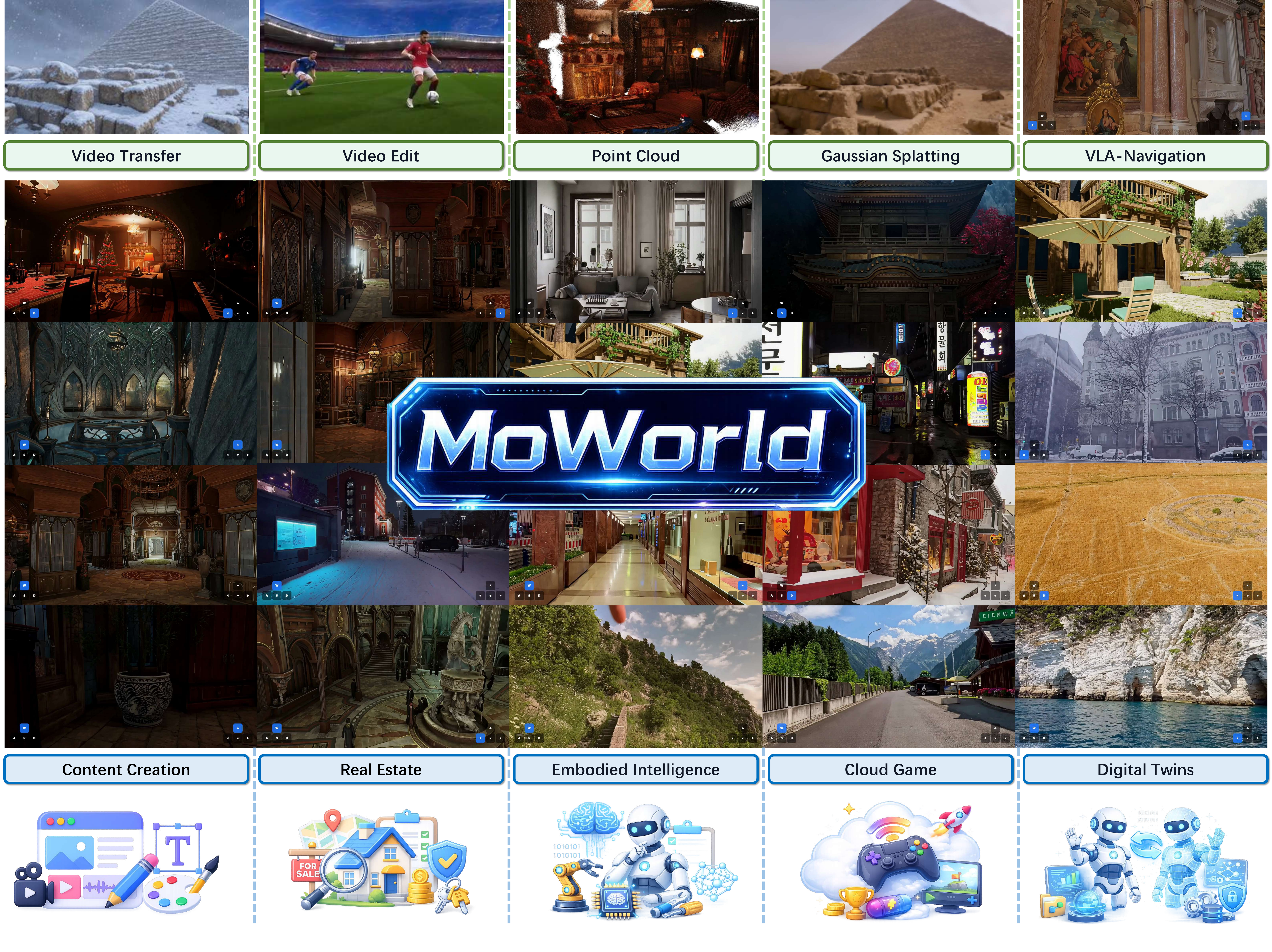}
  \vspace{-8mm}
  \caption{Overview of MoWorld applications across diverse downstream tasks.}
  \label{fig:teaser}
\end{center}
\end{figure}

Foundation Models~\cite{agpt,vggt,wang2026vggt,seedance,longcat,dens3r,hunyuanvideo,qwen3,veo31,kling25turbo,sora2,hunyuanvideo2025,ren2025cosmos,valevski2024diffusion} have rapidly transformed the landscape of artificial intelligence, driving an unprecedented leap toward more general and capable intelligent systems. Following the remarkable success of Large Language Models (LLMs)~\cite{qwen3,deepseek,glm,deepseekv3} and Multimodal Large Language Models (MLLMs)~\cite{qwen3-vl,seed-vl,gemini,radford2021learning,hunyuan-gamecraft,pan}, the next frontier is shifting from understanding and generating content to modeling the physical world itself. In this context, World Models are emerging as a new generation of foundation models that enable intelligent agents to perceive, reason about, and interact with dynamic environments~\cite{genie3,lingbot-world,hyworld15,sun2025worldplay,mao2025yume15,ali2025world,zhen2025tesseract,worldcanvas,ye2026world,bi2026motus,li2026world,song2026learning,cosmos,long2025survey}. Unlike conventional foundation models, which focus primarily on understanding and generating observations, World Models shift the modeling objective from observations to the world itself. Their goal is not merely to synthesize realistic images or videos, but to learn the underlying dynamics that govern how the physical world evolves under environmental changes and embodied agent actions. Equipped with such an internal model of the world, intelligent agents can predict future states, simulate interactions, reason over alternative outcomes, and plan long-horizon behaviors, making World Models a fundamental element of general embodied intelligence.

Rather than pursuing ever-larger World Models, we argue that the next stage of the field lies in making them practical. Future World Models must jointly optimize model capability, computational efficiency, deployment cost, and real-time inference performance, enabling scalable adoption beyond research laboratories into real-world intelligent systems. In particular, high-frame-rate inference is essential for latency-sensitive closed-loop applications such as embodied intelligence and autonomous driving, where slow generation fundamentally limits interaction, planning, and control. 

We hereby define \textbf{Flash World Models} as a new class of real-time world models capable of sustaining generation at 30 FPS (frames per second) or higher, enabling responsive interaction with dynamic environments. In computer graphics and interactive media, 30 FPS is widely recognized as the minimum threshold for achieving real-time rendering and delivering a fluid interactive experience. The seminal textbook Real-Time Rendering states real-time graphics—video games—"typically produce more than 30 frames per second (fps)" —at these speeds the player is fully focused on actions and reactions within the virtual environment. However, we find that existing popular solutions for world models \cite{genie3,lingbot-world,he2025matrix,he2025matrix} are hard to satisfy the criteria. 

In this paper, we present \textbf{MoWorld}, a practical \textbf{Flash World Model} designed through holistic data--algorithm--system--hardware co-design. Rather than optimizing a single component of the pipeline, MoWorld establishes an end-to-end framework spanning scalable data generation, efficient model training, accelerated inference, and deployment-oriented system optimization. Every stage is jointly optimized to maximize throughput while preserving generation quality, \textbf{enabling real-time world simulation at up to 50 FPS without the need of high-end GPUs}. As a result, MoWorld achieves strong generative performance while substantially reducing both training and inference costs.

To realize this vision, MoWorld is built upon four tightly coupled pillars that span the entire lifecycle of World Models.

First, we tackle the data bottleneck by building upon our research on Visual Geometry Transformers and large-scale geometry foundation models~\cite{vggt,vggsfm,wang2026vggt,liu2026streamcachevggt,chen2026hd}. Leveraging these geometry-centric representations, we develop a scalable data engine that curate the large-scale, high-quality World Model training data with accurate camera pose. Coupled with geometry-aware quality assessment and trajectory verification, the resulting corpus spans diverse and high-quality indoor and outdoor environments, providing the foundation for World Model pre-training.

Second, we introduce a progressive spatial-temporal capability learning paradigm that enables efficient scaling of World Models. Rather than directly optimizing long-context generation, the model progressively acquires increasingly sophisticated capabilities, from local interaction and camera response to long-horizon spatial memory and global world-state consistency. Combined with memory-efficient training strategies for ultra-long videos, this paradigm substantially improves scalability while preserving visual fidelity, camera controllability, and temporal consistency.

Third, we develop an efficient denoising-step distillation framework together with a series of optimization techniques spanning pre-training, distillation, and inference, significantly reducing both training and sampling cost while maintaining strong generative performance.

Finally, we extend optimization beyond model design through a deployment-oriented algorithm--system--hardware co-design framework. By jointly optimizing across the pipeline, parallelism, and kernel levels, including mixed-precision computation, dynamic quantization, parallel execution, and hardware-aware scheduling, MoWorld achieves real-time world simulation at up to 50 FPS with proper computational resources while reducing end-to-end deployment cost by 30--50\% in practical settings.

MoWorld is the first real-time interactive World Model designed for the Neural Processing Units (NPUs). Through operator-level optimization, compiler-aware scheduling, and NPU-native execution pipelines, MoWorld enables real-time inference at up to 50 FPS directly on widely available NPUs, eliminating the need for dedicated high-end GPUs. As NPUs are already integrated into many modern edge devices and intelligent systems, this substantially lowers deployment cost, power consumption, and hardware barriers, making practical World Models significantly more accessible for real-world applications.

By jointly integrating high performance and low cost, MoWorld establishes a unified world modeling framework that balances effectiveness and efficiency. We conduct systematic evaluations of MoWorld on standard image-to-video quality benchmarks and camera-controllable world model generation benchmarks. Experimental results show that MoWorld achieves leading performance across multiple dimensions, including subject consistency, background consistency, aesthetic quality, image preservation, and camera controllability, among others. It demonstrates that MoWorld not only remains competitive in general video generation quality, but also provides stable camera control and long-horizon consistency for interactive world modeling. As shown in Figure~\ref{fig:teaser}, MoWorld serves as a unified world model that enables a broad range of downstream applications across diverse domains. 

%% file: sections_en/2-data-engine.tex
\section{Geometric Aware Data Engine}

Data is the most critical asset for scaling World Models. Prior scaling studies have consistently shown that both data scale and data quality jointly determine model capability \cite{wang2026vggt}. In MoWorld, we argue that high-quality world modeling requires data that explicitly links visual appearance, camera motion, and controllable signals, rather than loosely paired video-text clips commonly used in conventional video generation. Based on team`s prior experience in 3D vision and visual geometry transformer \cite{liu2026streamcachevggt, chen2026hd}, we hereby use a geometric aware data engine to produce the data.

\begin{figure}[t]
\begin{center}
  \includegraphics[alt={},width=\textwidth]{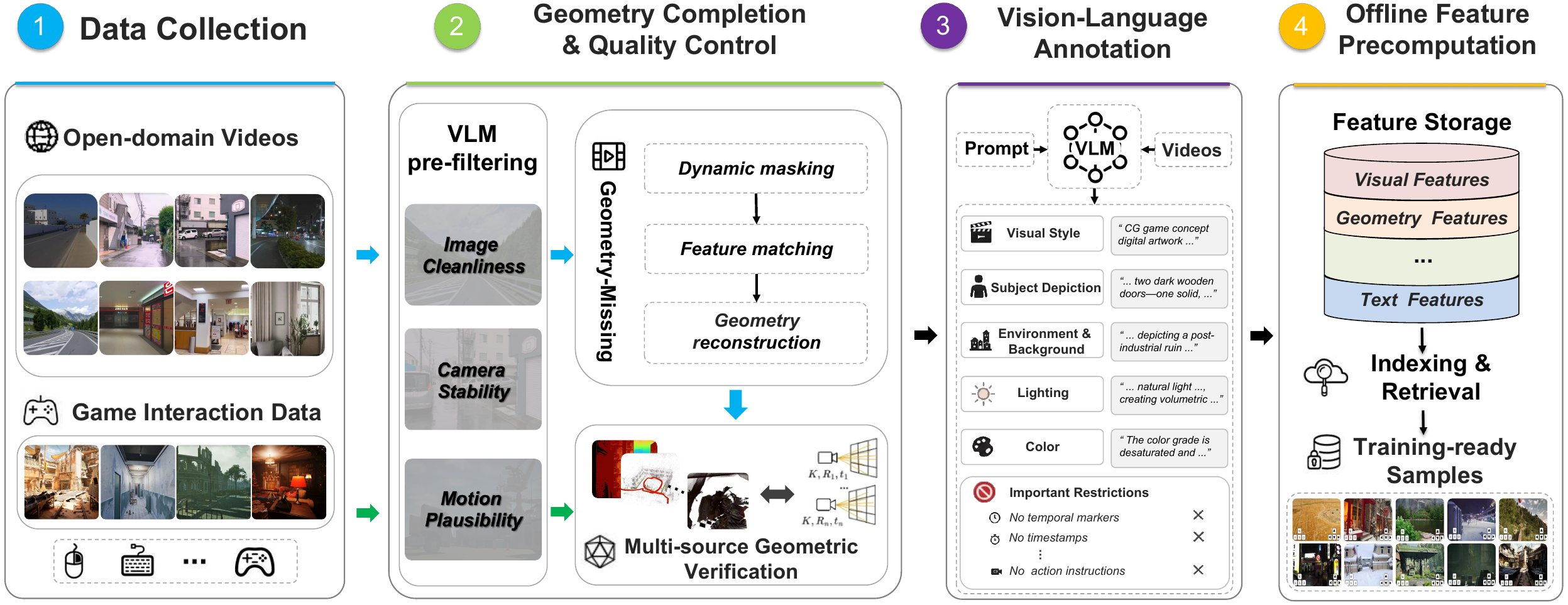}
  \caption{Overview of the MoWorld data engine. Multi-source videos are converted into training-ready samples through geometry completion, data quality control, vision-language model annotation, and precomputed caching.}
  \label{fig:data-engine}
\end{center}
\end{figure}

Specifically, each training sample must encode temporally aligned camera geometry, scene-level semantic context, and precomputed representations for efficient large-scale learning. As shown in Fig.~\ref{fig:data-engine}, the pipeline contains four stages. \textit{(1)} Data collection provides two complementary sources: open-domain videos for visual coverage and game-scene data for explicit control and geometry. \textit{(2)} Geometry completion and quality control recover missing camera conditions and reject samples with unreliable geometry or visual content. \textit{(3)} Vision-language model annotation assigns scene-level descriptions while keeping camera motion out of the text condition. \textit{(4)} Precomputed caching stores the processed modalities in a format that can be directly consumed during large-scale training.

Our protocol shares the design of VGGT-Omega \cite{wang2026vggt} geometry-centric data construction, Beyond camera pose supervision, we further annotate dense 3D point clouds for each scene, enabling downstream spatial reasoning, reconstruction, and world-state modeling tasks.

MoWorld is built entirely on proprietary data acquisition and in-house processing pipelines from up to 500 annotators, including both real-world captures and large-scale synthetic generation. This ensures strict control over data quality, geometric consistency, and distribution design, forming a clean and scalable foundation for World Model pre-training.

%% file: sections_en/3-pretrain-english.tex
\section{Model Pretraining}

\begin{figure}[t]
  \centering
  \includegraphics[alt={},width=0.8\textwidth]{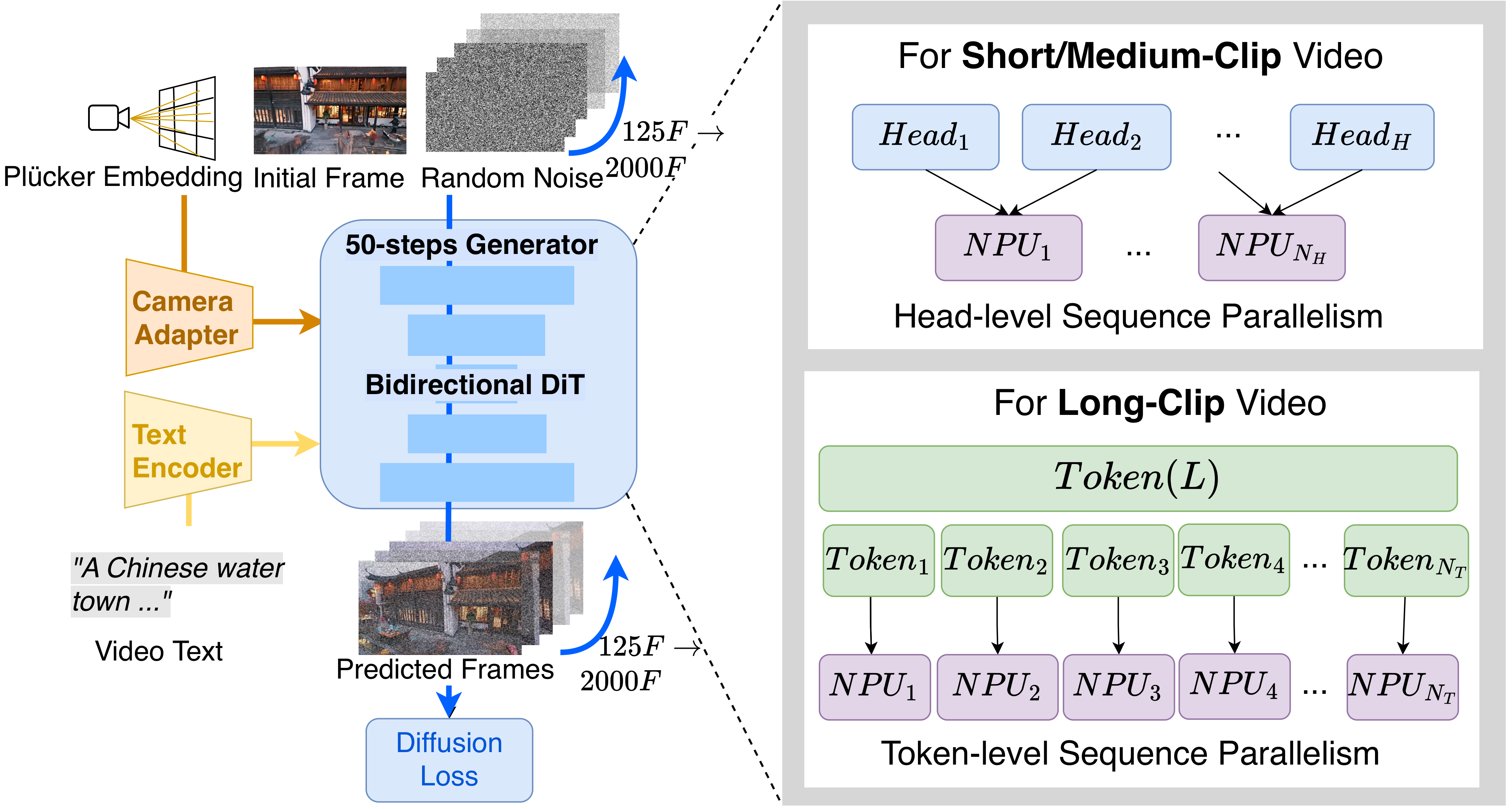}
  \caption{MoWorld is pretrained on NPU clusters with sequence parallelism (SP), distributing attention heads or input tokens across devices for scalable long-sequence learning. The model is jointly conditioned on the initial frame, text, and camera geometry, enabling progressive scaling of video generation from 125 to 2000 frames. $N$: number of NPU groups at different stages, with ($N_H \ll N_T$).
  }
  \label{fig:pretrain-overview}
\end{figure}

\subsection{Overview}\label{subsec:pretrain-overview}
The pretraining stage of MoWorld extends general video generation models to a camera-controlled, long-horizon generative framework for interactive world modeling. The model is conditioned on an initial frame, text prompts, and camera trajectory signals, and is required to faithfully follow these controls while generating high-fidelity video with long-horizon temporal consistency in scene structure, viewpoint transitions, and motion dynamics.
We conduct large-scale pretraining on camera-controlled data using a NPU cluster. This pretraining process evolves the model into a strong controllable world-model foundation for subsequent distillation and real-time inference.

We define MoWorld as $f_{\theta}$, which takes an initial frame $\mathbf{x}_0$ as visual initialization, text input $\mathbf{c}^{\text{txt}}$ as semantic context, and a sequence of camera control signals $\mathbf{c}^{1:T}_{\text{cam}}$ over timesteps $1:T$. The model is trained to predict future frames $\hat{\mathbf{x}}_{1:T}$ that align with the ground-truth (GT) sequence $\mathbf{x}_{1:T}$.
Following standard diffusion training \cite{ho2020denoising,rombach2022high,blattmann2023align,scalable-diffusion}, we condition on $\mathbf{x}_0$, $\mathbf{c}^{\text{txt}}$, and $\mathbf{c}^{1:T}_{\text{cam}}$. At diffusion timestep $t$, the model receives a noisy latent representation $\mathbf{z}_t$ of the future frames, with $t$ controlling the noise level, and learns to denoise it.
The pretraining objective is:
\begin{equation}
\mathcal{L}_{\mathrm{pre}}(\theta)=
\mathbb{E}_{\mathbf{x}_{0:T},\,\mathbf{c}^{\text{txt}},\,\mathbf{c}^{1:T}_{\text{cam}},\,t}
\left[
\left\|
f_{\theta}\!\left(\mathbf{x}_0,\mathbf{c}^{\text{txt}},\mathbf{c}^{1:T}_{\text{cam}}, t, \mathbf{z}_t\right)
-\mathbf{v}_t
\right\|_2^2
\right]
\label{eq:pretrain-objective}
\end{equation}
Here, $\mathbf{v}_t$ denotes the target of the diffusion parameterization derived from GT video future sequence (e.g., velocity prediction). 
At inference time, the model is conditioned on the first frame, text, and camera trajectory, and iteratively denoises a random latent to generate future frames. This pretrained backbone serves as the teacher model for subsequent distillation.

Architecturally, MoWorld follows the classical high-/low-noise MoE~\cite{moe,MoRE2025} design \cite{wan}, with an additional camera adapter to model camera control signals. Where, high-noise experts model global structure, layout, and large-scale motion, while low-noise experts refine fine details, including texture, edges, appearance consistency, and local temporal coherence. This model structure improves controllable video synthesis and provides a foundation for later long-video training, causal distillation, and efficient low-latency inference. 

An overview of the pretraining pipeline is illustrated in the Fig.~\ref{fig:pretrain-overview}. Our pretraining consists of three components, detailed in the following sections: (i) camera control injection into visual tokens to enable controllability (Sec.~\ref{subsec:camera-control}); (ii) curriculum training that progressively extends temporal horizons from short to long sequences (Sec.~\ref{subsec:training-curriculum}); and (iii) efficient large scale training system on NPUs (Sec.~\ref{subsec:ascend-training}).

\subsection{Cameral Control}\label{subsec:camera-control}

Camera control~\cite{cameractrl,zheng2024cami2v,bahmani2024vd3d,xu2024camco,wang2024cpa,liu2026camgeo} is a key conditioning signal in MoWorld for constructing an interactive world model, ensuring that generated frames faithfully follow future camera trajectories while maintaining viewpoint, spatial layout, and motion consistency. To achieve this, per-frame camera intrinsics and extrinsics are converted into dense geometric representations and injected into the visual token at the patch-token level.

The camera geometric condition $\mathbf{c}^{1:T}_{\text{cam}}$ is constructed from per-frame intrinsics and extrinsics. The intrinsic matrix $\mathbf{K}_i$ and the camera-to-world transformation $\mathbf{T}^{(i)}_{c2w}$ of the $i$-th frame be defined as:
\begin{equation}
\mathbf{K}_i =
\begin{bmatrix}
f_x & 0 & c_x\\
0 & f_y & c_y\\
0 & 0 & 1
\end{bmatrix},
\qquad
\mathbf{T}^{(i)}_{c2w}=[\mathbf{R}_i|\mathbf{t}_i].
\label{eq:camera-pose}
\end{equation}
For a pixel $\mathbf{u} = (u, v, 1)^\top$, we first compute its normalized ray $\mathbf{d}_c$ in the camera coordinate system using the intrinsic matrix, then transform it to the world coordinate ray $\mathbf{d}_w$ via the extrinsic parameters, and finally obtain the corresponding moment vector $\mathbf{m}_w$ in the world coordinate system.
\begin{equation}
\mathbf{d}_{c}=
\frac{\mathbf{K}_i^{-1}\mathbf{u}}
{\|\mathbf{K}_i^{-1}\mathbf{u}\|_2},
\qquad
\mathbf{d}_{w}=\mathbf{R}_i\mathbf{d}_{c},
\qquad
\mathbf{m}_{w}=\mathbf{t}_i \times \mathbf{d}_{w}
\label{eq:camera-ray}
\end{equation}
To jointly encode ray direction and spatial position in the camera condition, we represent 3D rays using Plücker coordinates \cite{cameractrl,seva} as the combination of the world-coordinate ray direction and its corresponding moment vector:
\begin{equation}
\mathbf{r}_{i,u,v} =
\begin{bmatrix}
\mathbf{m}_{w} \\
\mathbf{d}_{w}
\end{bmatrix}
\in \mathbb{R}^{6},
\qquad
\mathbf{R}_{\text{cam}} = \left\{ \mathbf{r}_{i,u,v} \right\}_{i,u,v=1}^{T,H,W} \in \mathbb{R}^{T \times H \times W \times 6}.
\label{eq:plucker-ray}
\end{equation}
Here, $\mathbf{R}_{\text{cam}}$ denotes the per-frame camera conditioning matrix aggregated over frames and pixels. A Camera Adapter $A_\phi$ projects $\mathbf{R}_{\text{cam}}$ into the visual token space, producing $N$ latent patch tokens with hidden dimension $d$. The camera-conditioned visual input at diffusion step $t$ is then formed by combining the patch embedding of the noisy latent $\mathbf{z}_t$ and the camera tokens $\mathbf{C}_{cam}$:
\begin{equation}
\mathbf{C}_{\text{cam}} = A_{\phi}(\mathbf{R}_{\text{cam}}) \in \mathbb{R}^{N \times d}, 
\qquad
\mathbf{X}_{\text{cam}}^{(t)} =
\mathrm{PatchEmbed}(\mathbf{z}_t) + \mathbf{C}_{\text{cam}}.
\label{eq:camera-token-injection}
\end{equation}
The key advantage of this design is that camera geometry is directly injected at the patch-token level, enabling the visual model to incorporate viewpoint constraints when modeling each local spatiotemporal block, thereby tightly coupling geometric consistency with the video generation process.

\subsection{Curriculum Cross-Frame Training}\label{subsec:training-curriculum}

MoWorld adopts a curriculum cross-frame learning strategy during pretraining, progressively increasing video sequence length to improve visual generation, camera controllability, and long-context modeling, while maintaining low pretraining cost. Training progresses from low-cost short clips, focusing on visual quality and camera controllability, to medium clips emphasizing temporal consistency, and finally to long clips targeting spatial memory and global world-state preservation. Most training is conducted in the low-cost short/medium-clip regime, with long clips introduced only in the final stage, yielding improved visual quality, camera control, and long-horizon stability at reduced training cost.

Specifically, Short-Clip (125F/250F, $T \in \{125, 250\}$) training focuses on establishing stable appearance generation, initial frame consistency, and basic camera control. In this stage, the model learns to respond to camera trajectories such as forward motion, turning, and orbiting within short horizons, while suppressing jitter and local structural distortion induced by control signals. Training then proceeds to Medium-Clip (500F/1000F, $T \in \{500, 1000\}$), where the extended temporal window introduces stronger constraints on object positioning, scene layout, and motion coherence, addressing the weak mid-range drift observed in short-clip training. In the Long-Clip (2000F, $T=2000$) stage, the objective shifts to long-context modeling, requiring the model to maintain spatial consistency during extended navigation, revisitation, and scene re-entry, preventing spatiotemporal feature drift and global geometric inconsistency across time. Each stage inherits weights from the previous one and reuses a unified cached data interface. This curriculum training yields a pretrained model that preserves the visual fidelity of the backbone while establishing a strong foundation for long-horizon video generation.

\subsection{Efficient Training System on NPUs}\label{subsec:ascend-training}
To enable large-scale NPU training from short- to long-horizon video generation, we focus on improving training efficiency. We first use a high-/low-noise resource pool assignment, together with offline caching to stabilize input throughput and improve data loading efficiency. We further enhance computation efficiency via optimized distributed parallelism across multiple NPUs. For Long-Clip video training (2000F), we adopt Sequence Parallelism (SP) \cite{fang2024usp,deepspeed} to enable efficient scaling to extreme sequence lengths. The details will be discussed in following subsections.

\subsubsection{Resource Organization and Input Caching}
Training resources follow the high-/low-noise MoE partitioning and are deployed on two NPU pools, ensuring alignment between model experts and hardware allocation.
On the input side, multimodal signals are offline encoded into cached video latents, text embeddings, and camera condition tensors. Training directly consumes these cached features, avoiding on-the-fly encoding, geometric recomputation, which become bottlenecks in long video training.
This cached pipeline removes host-side preprocessing overhead, stabilizes data throughput, and ensures efficient scaling under large scale curriculum training.

\subsubsection{Distributed Parallelism on NPUs}
\label{subsec:distributed_parallelism}
To enable efficient multi-NPU training, we adopt distributed parallelism with sharding of model parameters, gradients, and optimizer states using Fully Sharded Data Parallel (FSDP) \cite{rajbhandari2020zero}. Computation parallelism is more challenging, as video training is dominated by attention, whose memory scales with sequence length. As the number of frames (T) increases, the token sequence grows linearly with (T), while attention memory scales quadratically, leading to:
\begin{equation} L = \mathcal{O}(T), \qquad \mathrm{AttnMemory} = \mathcal{O}(T^2), \end{equation}
This quadratic scaling places significant pressure on model residency, activation memory, and HBM bandwidth. However, model parameter sharding alone is insufficient, as it does not mitigate attention kernel memory access, which becomes the primary bottleneck in long-sequence training.

To address this, MoWorld adopts Ulysses SP \cite{deepspeed}. It first partitions attention heads across ranks, assigning each device a subset of heads along with the corresponding query/key/value (QKV) views. Cross-rank communication is performed via HCCL to exchange necessary attention states.
Given this partition, each rank then computes attention locally on its assigned heads using fusion NPU attention kernel through the CANN runtime. The kernel performs QKV projection, softmax, and output aggregation in a fused, block-wise manner without explicitly materializing the full L×L attention matrix, thereby reducing HBM memory traffic and enabling efficient long-sequence processing.

\subsubsection{Sequence Parallelism for Long-Clip Video Training}

For long-clip settings (e.g., $T=2000$), Ulysses SP is insufficient, as its partitioning is restricted to the attention-head dimension, yielding fixed parallelism that does not scale with sequence length. This limits flexibility and efficiency under extreme video lengths, where attention complexity grows rapidly with $L$.

To address this, we adopt Unified Sequence Parallelism (USP), which generalizes sequence parallelism to token-level partitioning. Specifically, the sequence length $L$ is distributed across ranks, enabling finer-grained and scalable workload balancing as $L$ increases with $T$. This design provides more flexible parallelism and improves efficiency for extreme long-context video generation.

%% file: sections_en/4-distillation-english-simple.tex
\section{Autoregressive Distillation}

\begin{figure}[t]
  \centering
  \includegraphics[alt={},width=0.76\textwidth]{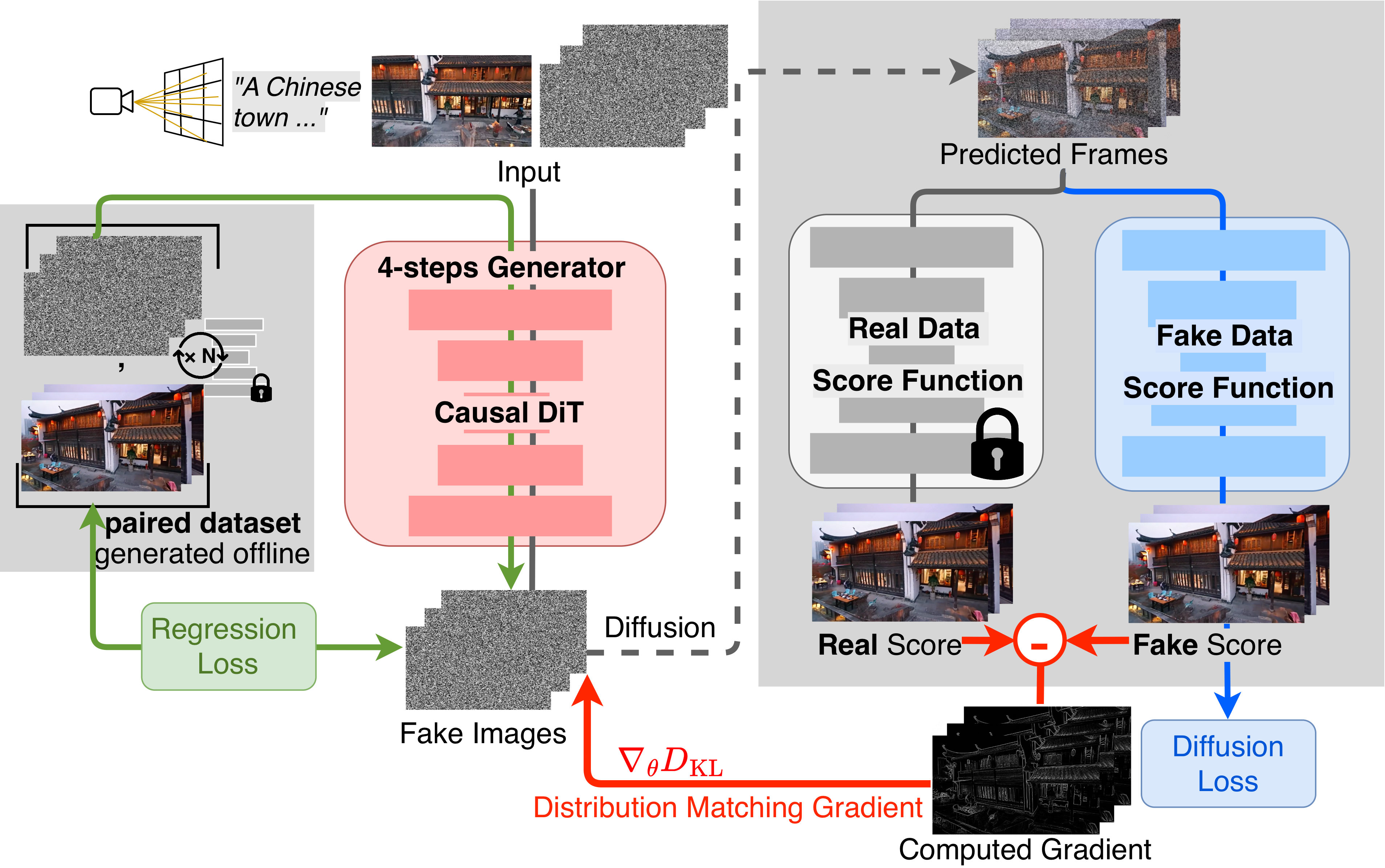}
  \vspace{5mm}
  \caption{
  In MoWorld distillation stage, a unified latent-space distillation objective is applied to compress inference steps from 50 to 4, while reformulating the bidirectional DiT architecture into a self-driven causal DiT for efficient autoregressive generation.
  }
  \label{fig:distillation-overview}
\end{figure}

\subsection{Overview}

MoWorld's distillation stage converts a pretrained teacher model with bidirectional window modeling capability into a few-step autoregressive (AR) student model for real-time interaction. The teacher model provides a high-quality distribution prior over a complete video window, whereas the student model generates future content by rolling forward in time during inference~\cite{self-forcing,self-forcing++,dmd,yin2024improved,causvid,distillation-hinton,magi}. The student is required to preserve visual quality, camera responsiveness, and long-range consistency under limited context and only a small number of sampling steps. Distillation serves two purposes: (1) Few-step Generation, compressing multi-step diffusion sampling into few-step inference, and (2) Autoregressive Generation, transferring bidirectional window modeling into a continuously executable AR process.

To achieve this, MoWorld decomposes AR distillation into three components. First, we introduce a history context selection module that maintains a compact set of latent frames instead of all past tokens or a full KV cache, thereby bounding memory growth (Sec.~\ref{subsec:distill-history}). Second, AR Flow Matching pretraining adapts bidirectional pretrained weights to a causal, chunk-wise interface via ground-truth prefixes and current-chunk supervision, stabilizing early AR generation without requiring ODE-based initialization or multi-step teacher sampling \cite{causalforce} (Sec.~\ref{subsec:autoregressive}). Third, Self-Forcing distillation trains the student on its own AR rollouts, where a frozen teacher and a fake model provide full-video distributional feedback, enabling correction of rollout errors while compressing generation into few-step inference (Sec.~\ref{subsec:autoregressive}).

\subsection{History Context Selection Design}
\label{subsec:distill-history}

Historical information grows continuously during autoregressive generation. Feeding all previous latent frames into the current window would make the attention length and memory cost increase over time, while relying only on the most recent segment would cause the model to lose early appearance, scene layout, and revisited viewpoints in long videos~\cite{xiao2026worldmem,gao2024cat3d,liu2026reconx,ren2025gen3c}. MoWorld therefore redesigns the History Context selection strategy. From all previously generated latent frames, it selects recent frames to maintain short-range continuity, early frames to stabilize subject appearance and global layout, and camera-related frames to handle revisiting, orbiting, or looking back from similar viewpoints.

For a non-initial chunk, MoWorld selects the most recent latent frames together with the initial frame of the inference sequence, providing both clean reference appearance and recent temporal context for future generation.
Camera-related History is retrieved using camera latent vectors. For the current chunk, the system first computes the mean camera latent of that chunk and then compares it with the camera latent associated with each candidate historical frame. A smaller distance indicates a closer viewpoint or motion state and is selected as current chunk's camera-related frames. 
For the initial chunk, all latent frames are initialized as the first frame of the inference sequence.
Then Each completed chunk is written back to the History Bank. The bank stores latent frames rather than KV Cache, which substantially reduces the persistent memory footprint. 
Compared with using only the most recent history, this retrieval strategy can reintroduce relevant scene states when a viewpoint reappears, thereby reducing drift in geometric structure and subject identity.

During inference, the selected History latents are concatenated with the current chunk along the temporal dimension and jointly fed into the denoising model. RoPE~\cite{rope} uses global latent time IDs instead of restarting the positional indices within each chunk. The model finally keeps only the output tokens corresponding to the current chunk; History tokens are not prediction targets and are not updated by the current chunk. This design allows historical information to participate in current generation without maintaining a long-lived KV Cache, while keeping the History context fixed across multiple denoising steps.

\subsection{Autoregressive Flow Matching Pretraining}
\label{subsec:autoregressive}
The pretrained bidirectional model cannot be directly used as an autoregressive generator~\cite{self-forcing,self-forcing++,dmd,causalforce,yin2024improved,causvid}. It is trained to denoise a full video window with bidirectional context, whereas AR inference conditions only on causal inputs, including history context, an anchor frame (the initial frame of the current chunk), future camera conditions, and the noisy latent of the current chunk. Without adaptation, the model cannot reliably interpret this new token layout, leading to unstable chunk generation and poor rollout behavior. This would also make the subsequent Self-Forcing distillation less reliable, since the Fake Model and the Teacher--Fake prediction gap would be computed from invalid student samples.

We therefore introduce AR Flow Matching pretraining as an intermediate adaptation stage before distillation. This stage has three roles. First, it constructs the prefix history and boundary anchor from ground-truth latents, allowing the student to learn history reading and cross-chunk boundary alignment under stable historical conditions. Second, it applies Flow Matching supervision only to the current chunk, adapting the bidirectional pretrained weights to the causal chunk-wise interface while preserving the original denoising capability. Third, it avoids the ODE-initialization or multi-step teacher-sampling stage used in some distillation pipelines, and instead relies on cached ground-truth latents and camera conditions. As a result, AR Flow Matching pretraining improves the stability of early autoregressive generation while reducing the computational overhead before Self-Forcing distillation.

\subsubsection{Ground-Truth Prefix History Construction}

During training, the student unfolds each video into fixed-length latent windows, referred to as AR chunks. We denote $\mathbf{X}_i$ as the sequence of latent frames corresponding to the $i$-th AR chunk.
In training, a current chunk index is uniformly sampled as
\begin{equation}
i \sim \mathrm{Uniform}\{0, \ldots, i_{\max}\},
\end{equation}
where $i_{\max} = \left\lfloor \frac{L_{\text{video}}}{L_{\text{chunk}}} \right\rfloor - 1$ denotes the last valid chunk index. This ensures that $i$ spans all valid chunks from the first to the final chunk supported by the video length.

For the history preceding the current chunk, the system does not rely on student-generated outputs. Instead, the History Bank is constructed directly from ground-truth latent frames:
\begin{equation}
\mathcal{M}_{<i}=
{(\mathbf{x}_j, \mathbf{c}^{(j)}_{\mathrm{cam}})\mid j \in \mathcal{I}_{<i}},
\end{equation}
where $\mathcal{I}_{<i}$ denotes the set of frame-level indices belonging to all chunks before the $i$-th chunk. Here, $j$ indexes frames (as opposed to $i$, which indexes chunks), $\mathbf{x}_j$ is the ground-truth latent frame, and $\mathbf{c}^{(j)}_{\mathrm{cam}}$ is its associated camera control signal.

If the current chunk is not the first, its anchor frame is defined as the last latent frame of the preceding ground-truth chunks:
\begin{equation}
\mathbf{X}_i^{\text{anchor}} = \mathbf{x}_{\max(\mathcal{I}_{<i})}.
\end{equation}

This teacher-forced prefix design follows the observation that clean historical context provides a stable autoregressive training signal~\cite{causalforce,yuan2026helios}. In our setting, it is instantiated as ground-truth History construction for chunk-level AR Flow Matching: the student predicts the current chunk under correct historical latents and camera poses, preventing early student errors from recursively contaminating History. As a result, the model can learn History reading and boundary alignment from stable contexts before handling self-generated histories during distillation.

\subsubsection{Autogressive Flow Matching Supervision}

Conditioned on the anchor frame $\mathbf{X}_i^{\text{anchor}}$, text prompt $\mathbf{c}^{\mathrm{txt}}$, current chunk camera condition $\mathbf{C}^{\mathrm{cam}}_{i}$, ground-truth prefix history $\mathcal{M}_{<i}$, diffusion timestep $t$, and the corresponding noisy latent $\mathbf{Z}^t_{i}$ of the current chunk, the autoregressive student training objective is:
\begin{equation}
\mathcal{L}_{\mathrm{FM}}(\theta)=
\mathbb{E}_{\mathbf{X}_{i},\,\mathbf{c}^{\text{txt}},\,\mathbf{C}_{i}^{\text{cam}},\mathcal{M}_{<i},\,t}
\left[
\left\|
f_{\theta}^{\mathrm{AR}}
\left(
\mathbf{X}_i^{\text{anchor}},
\mathbf{c}^{\mathrm{txt}},
\mathbf{C}^{\mathrm{cam}}_i,
\mathcal{M}_{<i},
t,
\mathbf{Z}^t_{i}
\right)
-\mathbf{V}^t_{i}
\right\|_2^2
\right]
\label{eq:pretrain-objective}
\end{equation}
where $\mathbf{V}^t_{i}$ denotes the flow target of current chunk.

In addition, we constrain the overlapped latent at the chunk boundary with an auxiliary anchor loss to stabilize frame continuity when the AR model generates chunk by chunk:
\begin{equation}
  \mathcal{L}_{\mathrm{anchor}}
  =
  \left\|
    \hat{\mathbf{X}}_{i}^{0}
    -
    \mathbf{X}_i^\text{anchor}
  \right\|_2^2.
\end{equation}
Here, $\hat{\mathbf{X}}_{i}^{0}$ denotes the predicted first latent frame of the current chunk. The supervised pretraining loss is
\begin{equation}
  \mathcal{L}_{\mathrm{AR}}
  =
  \mathcal{L}_{\mathrm{FM}}
  +
  \lambda_{\mathrm{anchor}}\mathcal{L}_{\mathrm{anchor}},
  \qquad
  \lambda_{\mathrm{anchor}}=1.0.
\end{equation}

\subsection{ODE-Initialization-Free Distillation Design}
\label{subsec:ode-initialization}
Some distillation methods~\cite{causvid} initialize the student with teacher ODE trajectories~\cite{causalforce} or multi-step teacher samples, which introduces a major efficiency bottleneck. The teacher contains separate high-noise and low-noise DiT experts, and generating multi-step teacher trajectories for long sequence would require repeated expensive rollouts before distillation even begins. As a result, the overall training throughput would be severely limited by teacher sampling.

We use the auto-regressive flow matching pretraining~\cite{flowmatch} avoids this initialization stage. It directly uses cached ground-truth latents and camera conditions to adapt the student to the autoregressive input distribution, without invoking multi-step teacher rollouts. This design substantially reduces the computational cost of distillation while still providing the student with stable chunk generation, History reading, and boundary alignment capabilities before Self-Forcing distillation starts.

We follow the distribution-matching idea of DMD-style methods~\cite{dmd}, but replace the student sampling path with the actual autoregressive rollout used at inference time as shown in Figure~\ref{fig:distillation-overview}. Given the initial frame $\mathbf{x}_0$, text condition $\mathbf{c}^{\mathrm{txt}}$, and chunk-level camera trajectories $\mathbf{C}^{\mathrm{cam}}$, the student generates a full latent video chunk by chunk:
\begin{equation}
  \hat{\mathbf{Z}}_{\theta}
  =
  \mathrm{Rollout}_{\theta}^{\mathrm{AR}}
  (\mathbf{x}_0, \mathbf{c}^{\mathrm{txt}}, \mathbf{C}^{\mathrm{cam}},\boldsymbol{\epsilon}; \mathcal{S}),
\end{equation}
where $\boldsymbol{\epsilon}$ is noise and $\mathcal{S}$ denotes the chunk schedule, stitching rule, and history context selection strategy. Since samples are produced through the same AR pipeline used during deployment, boundary errors, history-selection errors, and long-range drift are exposed during training.

Training alternates between a Fake Phase and a Student Phase. In the Fake Phase, the student and teacher models are frozen, and the Fake Model learns the Flow Matching velocity field of the current student-generated distribution. Given a student rollout $\hat{\mathbf{Z}}_{\theta}$, noise $\boldsymbol{\epsilon}$, and timestep $t$, we construct chunk-level noisy latents $\mathbf{Z}_t$ and target velocities $\mathbf{v}^{\mathrm{fake}}_t$ for Fake model.
\begin{equation}
  \mathbf{Z}_t
  =
  (1-t)\hat{\mathbf{Z}}_{\theta}
  +
  t\boldsymbol{\epsilon},
  \qquad
  \mathbf{v}^{\mathrm{fake}}_t
  =
  \boldsymbol{\epsilon}
  -
  \hat{\mathbf{Z}}_{\theta}.
\end{equation}
The Fake Model is optimized by
\begin{equation}
  \mathcal{L}_{\mathrm{fake}}
  =
  \mathbb{E}_{\mathbf{x}_0,\mathbf{c}^{txt},\mathbf{C}^{cam},t,\hat{\mathbf{Z}}_{\theta},\boldsymbol{\epsilon}}
  \left[
    \left\|
      f_{\phi}^{\mathrm{fake}}(\mathbf{x}_0,\mathbf{c}^{\mathrm{txt}},\mathbf{C}^{\mathrm{cam}},t,\mathbf{Z}_t)
      -
      \mathbf{v}^{\mathrm{fake}}_t
    \right\|_2^2
  \right],
\end{equation}
so that it approximates the velocity field induced by the student AR distribution. 

In the Student Phase, the teacher and Fake Model are frozen. The teacher estimates the velocity field of the real video distribution, while the Fake Model estimates that of the current student distribution on the same noisy latent $\mathbf{Z}_t$:
\begin{equation}
  \mathbf{v}^{\mathrm{real}}_t
  =
  f_{\theta}^{\mathrm{real}}(\mathbf{x}_0,\mathbf{c}^{\mathrm{txt}},\mathbf{C}^{\mathrm{cam}},t,\mathbf{Z}_t),
  \qquad
  \mathbf{v}^{\mathrm{fake}}_t
  =
  f_{\phi}^{\mathrm{fake}}(\mathbf{x}_0,\mathbf{c}^{\mathrm{txt}},\mathbf{C}^{\mathrm{cam}},t,\mathbf{Z}_t).
\end{equation}
Their difference defines the distribution-matching direction
\begin{equation}
  \mathbf{g}_{\mathrm{DMD}}
  =
  \mathbf{v}^{\mathrm{real}}_t
  -
  \mathbf{v}^{\mathrm{fake}}_t,
\end{equation}
which provides a proxy gradient that moves the AR student distribution toward the teacher distribution. The student objective is written as
\begin{equation}
  \mathcal{L}_{\mathrm{student}}
  =
  \mathbb{E}_{\mathbf{x}_0,\mathbf{c}^{txt},\mathbf{C}^{cam},t,\hat{\mathbf{Z}}_{\theta},\boldsymbol{\epsilon}}
  \left[
    \mathrm{sg}(\mathbf{g}_{\mathrm{DMD}})
    \cdot
    \hat{\mathbf{Z}}_{\theta}
  \right],
\end{equation}
where $\mathrm{sg}(\cdot)$ denotes stop-gradient. Thus, gradients are back-propagated only through the student rollout, while the teacher and Fake Model provide a full-video distribution-matching signal for the actual autoregressive generation chain. This enables efficient distillation from the teacher model to the student model for few-shot denoising.

%% file: sections_en/5-inference-english.tex
\section{Real-Time Inference}
\label{sec:inference}

\subsection{Overview}
\begin{figure}[t]
  \centering
  \includegraphics[alt={}, width=0.99\textwidth]{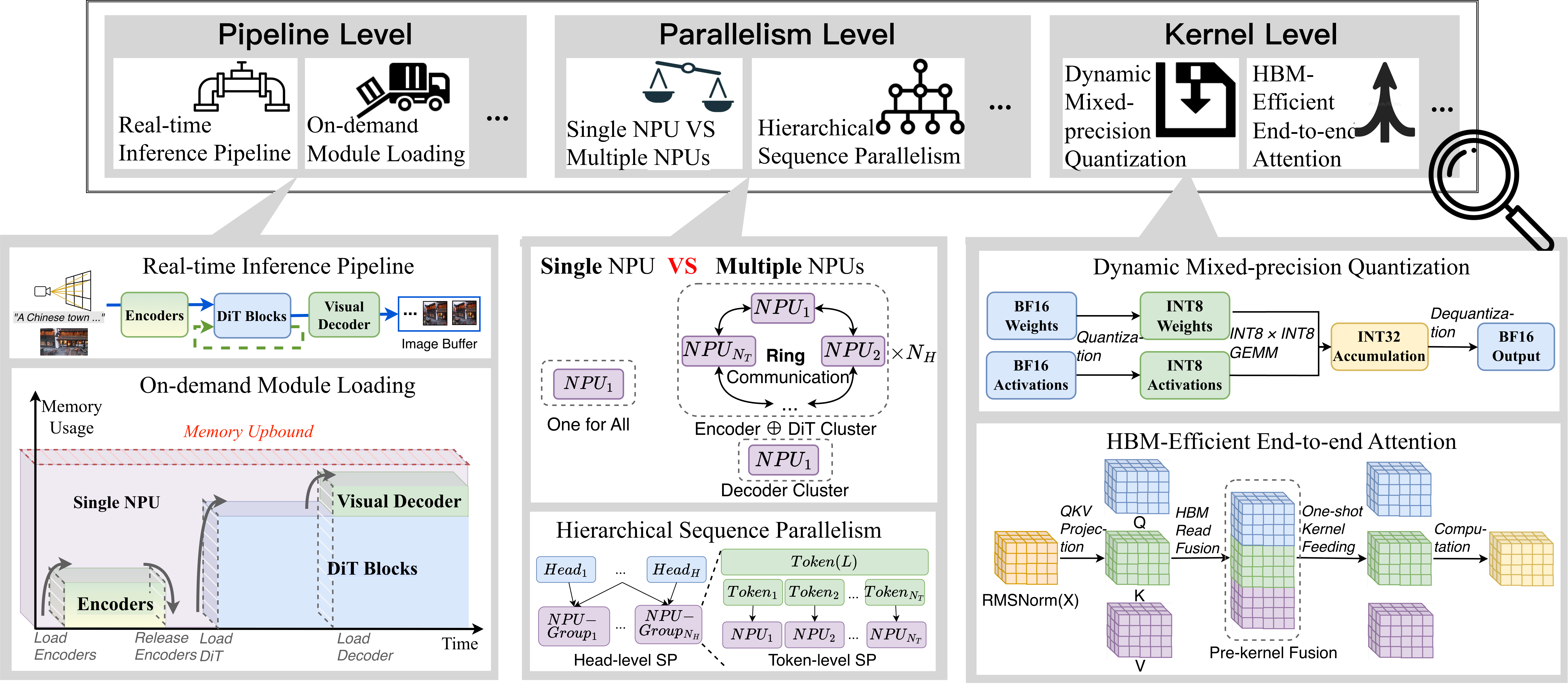}
  \caption{An overview of low-memory, low-latency real-time inference via pipeline-, parallelism-, and kernel-level optimizations.}
  \label{fig:inference-details}
\end{figure}

\begin{figure}[t]
  \centering
  \includegraphics[alt={},width=0.85\textwidth]{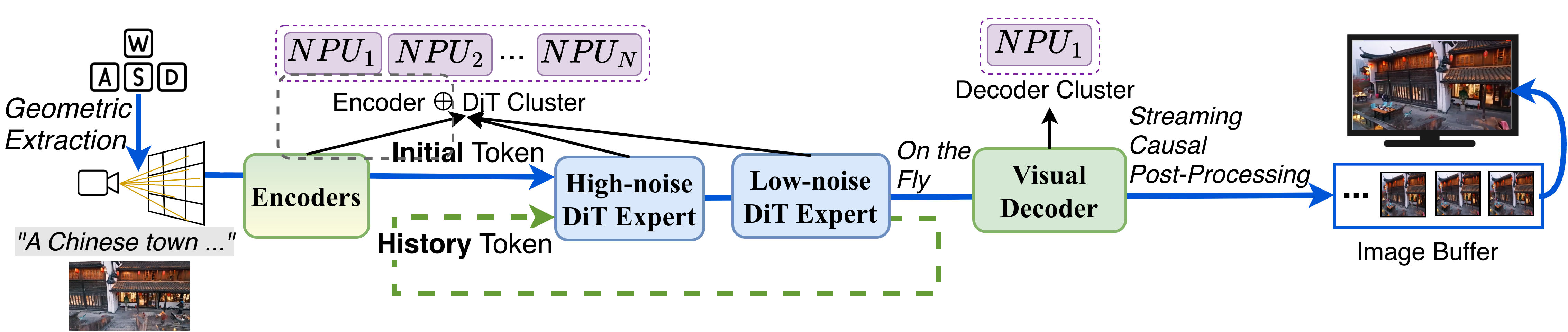}
  \caption{MoWorld real-time inference pipeline across multiple NPUs. The model performs autoregressive prediction conditioned on the initial frame, text, and user control commands, encoding and denoising them to generate future latent representations. The predictions are fed back as historical context for subsequent steps and decoded for display.
  }
  \label{fig:inference-overview}
\end{figure}

To enable real-time prompt human interaction with MoWorld under varying NPU resource regimes, the system must satisfy two constraints. (1) Low-memory Consumption under resource-constrained settings (e.g., single-NPU deployment), where all components—including modality-specific encoders and decoders, DiT experts, activations, communication buffers, and inference states—must fit within a strict memory budget to ensure feasibility under limited resources. (2) Low-latency Inference under both resource-constrained and resource-rich settings (e.g., single- or multi-NPU deployment), aiming for efficient parallel execution such that camera control signals are rapidly reflected in subsequent frames, while outputs are continuously decoded and streamed for real-time visualization.

To enable low-memory consumption, low-latency inference under varying NPU resource constraints, we address the problem at three hierarchical levels: (i) pipeline level, covering input encoding, model loading/unloading, and output generation (Section~\ref{subsec:pipeline-layer}); (ii) parallelism level, focusing on weight and token parallelization and their distribution across multiple sources (Section~\ref{subsec:parallelism-layer}); and (iii) kernel level, reducing effective computation via efficient operator designs (Section~\ref{subsec:kernel-layer}).

\subsection{Pipeline Level Optimization}
\label{subsec:pipeline-layer}
An overview and detailed illustration of the \textit{Real-time Inference Pipeline} are shown in the $x$: left and $y$: middle of the Figure~\ref{fig:inference-details} and Figure~\ref{fig:inference-overview}, respectively. Given an initial frame, video text, and user camera controls (W/A/S/D), we first convert the control inputs into Plücker embeddings via geometric extraction. Visual, textual, and camera embeddings are first extracted via dedicated encoders and used as the initial conditioning for the DiT backbone, which is autoregressively denoised by high- and low-noise experts sequentially to predict future-frame latent tokens. 
We adopt an autoregressive inference scheme in which previously generated frame latents are recycled as historical context for subsequent steps. Following the memory design in last section, we encode global temporal positions across the full generation sequence using rotary positional embeddings (RoPE). 
A VAE-based visual decoder then subsequently reconstructs raw frames from predicted latents. 
To reduce perceptual latency, multiple frames are generated per step and buffered for streaming display, enabling responsive execution of user commands with continuous output rendering.

In single NPU setting, memory is the primary bottleneck: the visual, text, and camera encoders, the DiT backbone, and the VAE decoder cannot be co-resident on a single NPU due to memory constraints, as shown in the $x$: left and $y$: bottom of Figure~\ref{fig:inference-details}.
 We therefore adopt a \textit{On-demand Module Loading} strategy: encoders are executed once at first step to produce the initial condition embeddings, after which their weights are released to reduce memory usage. The DiT and decoder are then loaded sequentially for generation. The resulting latent features are reused across autoregressive steps, eliminating the need for repeated encoder execution. By doing so, we eliminate redundant encoder memory footprint and allocate most of the available memory to the primary DiT blocks.

\subsection{Parallelism Level Optimization}
\label{subsec:parallelism-layer}
For multi-NPU deployment, parallel execution across NPUs is required to fully utilize available resources and reduce inference latency, making workload partitioning and inter-device communication critical challenges. As shown in Figure~\ref{fig:inference-overview} and the $x$: middle and $y$: middle of Figure~\ref{fig:inference-details}, we decouple the Encoder $\oplus$ DiT NPU placement from the decoder, deploying the visual decoder on a dedicated NPU and allocating the remaining NPUs to DiT computation, which dominates memory and compute cost; encoders are released after encoding the initial condition.

For resource allocation within the Encoder $\oplus$ DiT NPU cluster, we adopt a \textit{Hierarchical Sequence Parallelism} strategy, closely following but slightly modifying Section~\ref{subsec:distributed_parallelism} as shown in the $x$: middle and $y$: bottom of Figure~\ref{fig:inference-details}. First, attention heads are partitioned across NPUs (head-level parallelism). Within each head partition group, input tokens are further split via token-level sequence parallelism. This two-level decomposition reduces per-device memory pressure while maintaining high utilization. Compared to single-level partitioning, hierarchical splitting achieves lower end-to-end latency under the same hardware budget.

The inter-device communication within head partition groups is implemented via ring-based communication, while cross-group communication adopts standard AllToAll topology~\cite{ringattention,liu2025world}, as shown in the $x$: middle and $y$: middle of Figure~\ref{fig:inference-details}. It is then executed via HCCL collective operations, ensuring efficient exchange of KV states and partial activations across NPUs. By fully utilizing all available resources, the above design improves inference efficiency over single-NPU execution and enables real-time human–model interaction.

\subsection{Kernel Level Optimization}
\label{subsec:kernel-layer}
MoWorld employs high- and low-noise MoT models, where parameter duplication substantially increases memory overhead; thus, even with on-demand module loading, DiT blocks and the decoder cannot be co-resident on a single NPU. To further reduce memory pressure, we adopt \textit{Dynamic Mixed-Precision Quantization}, as shown in the right ($x$) and middle ($y$) of Figure~\ref{fig:inference-details}. Model weights and activations are initially loaded in BF16 and then converted via a one-step kernel warm-up quantization to INT8 precision, yielding approximately 2× memory reduction. INT8×INT8 accumulation is performed to produce INT32 intermediate results, followed by dequantization to recover BF16 outputs. This quantization scheme is applied only to the DiT blocks, while encoders remain in BF16 due to their sensitivity to precision and their transient execution under on-demand loading. This enables dynamic precision quantization, where activations are loaded and produced in BF16 while computation is performed in INT8 to reduce memory overhead.

To further improve NPU efficiency, we use an \textit{HBM-efficient End-to-end Attention} as shown in $x$: middle and $y$: bottom of Figure~\ref{fig:inference-details}. It first applies RMSNorm to stabilize the scale of activations with minimal computational overhead prior to QKV projection, ensuring numerically well-conditioned inputs while avoiding the costly mean-centering and variance computation required in standard LayerNorm. Unlike standard attention implementations, we adopt an NPU-fused attention kernel in which QKV projection and tiling are fused and transferred as a single HBM transaction, followed by on-kernel decomposition and online softmax computation. This reduces memory movement and improves throughput over conventional implementations, enabling competitive performance under tight NPU resources and load latency constraints.

%% file: sections_en/6-evaluation-english.tex
\section{Experiments}
\label{sec:evaluation}

\subsection{Quantitative Results}

We quantitatively evaluate MoWorld following the protocols commonly adopted by recent video-generation-based world model methods. Specifically, we conduct the evaluation using the official image-to-video benchmark data from the VBench-I2V branch~\cite{vbench}. Unlike the original VBench benchmark, which primarily targets text-to-video generation, VBench-I2V introduces an image-conditioned formulation in which a first-frame image is provided as input for each test sample and reports dedicated metrics for image-to-video models. It therefore evaluates how well a model preserves the initial visual state and is more suitable for assessing world models initialized from an input image.

We report eight metrics: SC (Subject Consistency), BC (Background Consistency), MS (Motion Smoothness), DD (Dynamic Degree), AQ (Aesthetic Quality), IQ (Imaging Quality), I2V-S (Image-to-Video Subject Consistency), and I2V-B (Image-to-Video Background Consistency). In particular, I2V-S and I2V-B are task-specific metrics introduced in VBench-I2V to measure the preservation of subject and background information from the input first frame, respectively. We additionally report a VBench-style Quality score, computed as a weighted average over the first six metrics, where all metrics have a weight of 1 except DD, which has a weight of 0.5. We also define Average as the arithmetic mean over all eight metrics, summarizing appearance consistency, motion quality, visual fidelity, and first-frame preservation.

In addition to the official VBench-I2V evaluation, we collect hundreds of test samples from the Internet. These samples are drawn from diverse sources, including AI-generated videos, game environments, and real-world videos, and cover a wide range of scenarios such as indoor scenes, natural environments, and in-the-wild settings. We extract the initial image, text description, and camera trajectory conditions from these videos to further evaluate the model under diverse visual scenarios and interactive camera-control settings.

The results in Table~\ref{tab:official_vbench} show that MoWorld achieves strong and balanced performance across these evaluations. MoWorld obtains the best aggregate Quality and Average scores, indicating that the proposed training and distillation pipeline preserves the general video generation capability of the foundation model while improving its robustness under world-model-style conditioning. More specifically, MoWorld performs strongly on SC and BC, suggesting stable subject identity and background structure. It also achieves competitive AQ and IQ scores, showing that camera-conditioned training does not substantially degrade visual aesthetics or imaging quality. Its strong I2V-S and I2V-B scores further demonstrate that MoWorld reliably preserves the subject and background information contained in the input first frame, which is essential for world models that treat a user-provided image as the initial state of the environment.

As shown in Table~\ref{tab:self_vbench}, the test samples collected from the Internet further highlight the advantage of MoWorld under interactive control. Compared with existing methods, MoWorld maintains high visual Quality and Average scores while incorporating explicit camera-trajectory conditions. This demonstrates that the model can generate viewpoint changes corresponding to the specified camera motion without substantially sacrificing subject consistency, background stability, or image-conditioned preservation.

\begin{table}[htbp]
\centering
\small
\caption{Results on a sampled subset of the official VBench-I2V dataset.
Quality is the weighted average over the first six dimensions, where DD
has a weight of 0.5 and all other dimensions have a weight of 1.
Average is the arithmetic mean over all eight dimensions.
Bold and underline indicate the best and second-best results, respectively.}
\label{tab:official_vbench}
\resizebox{\linewidth}{!}{%
\begin{tabular}{lcccccccccc}
\toprule
Model
& SC$\uparrow$
& BC$\uparrow$
& MS$\uparrow$
& DD$\uparrow$
& AQ$\uparrow$
& IQ$\uparrow$
& I2V-S$\uparrow$
& I2V-B$\uparrow$
& Quality$\uparrow$
& Average$\uparrow$ \\
\midrule

CameraCtrl \cite{cameractrl}
& \underline{95.44}
& \underline{95.36}
& \textbf{98.92}
& 38.00
& 54.86
& 66.27
& 95.95
& \underline{97.48}
& 78.15
& 80.29 \\

SEVA \cite{seva}
& 87.07
& 90.18
& 97.62
& \textbf{62.00}
& 52.65
& 61.40
& 93.83
& 95.64
& 76.35
& 80.05 \\

Lingbot \cite{lingbot-world}
& 93.91
& 94.94
& 97.80
& 46.00
& \underline{57.54}
& \textbf{71.20}
& \underline{97.27}
& 97.14
& \underline{79.71}
& \underline{81.97} \\

MoWorld
& \textbf{95.47}
& \textbf{95.68}
& \underline{98.13}
& \underline{60.00}
& \textbf{64.63}
& \underline{71.11}
& \textbf{98.04}
& \textbf{98.68}
& \textbf{82.73}
& \textbf{85.22} \\

\bottomrule
\end{tabular}}
\end{table}

\begin{figure}[!t]
  \centering
  \includegraphics[alt={},width=\textwidth]{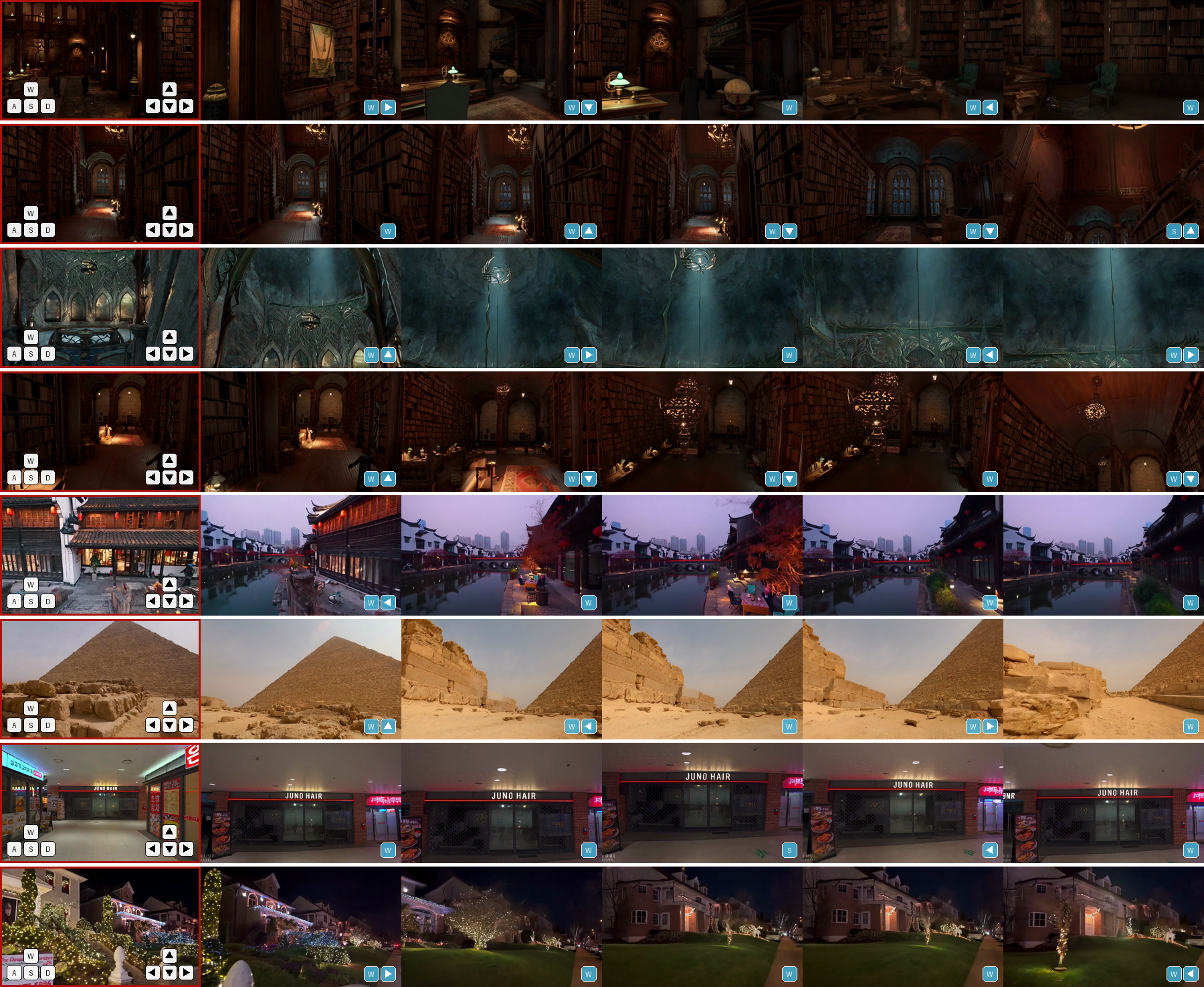}
  \caption{Qualitative results.}
  \label{fig:exp-visualization-1}
\end{figure}

\begin{table}[tbp]
\centering
\small
\caption{VBench-style eight-dimensional results on a sampled subset of our in-house camera/world-model dataset. Bold and underline indicate the best and second-best results, respectively.}
\label{tab:self_vbench}
\resizebox{\linewidth}{!}{%
\begin{tabular}{lcccccccccc}
\toprule
Model & SC$\uparrow$ & BC$\uparrow$ & MS$\uparrow$ & DD$\uparrow$ & AQ$\uparrow$ & IQ$\uparrow$ & I2V-S$\uparrow$ & I2V-B$\uparrow$ & Quality$\uparrow$ & Average$\uparrow$ \\
\midrule
SEVA \cite{seva} & 90.08 & 92.23 & \underline{98.48} & 56.00 & 53.53 & 50.09 & 96.40 & 93.18 & 74.98 & 78.75 \\
CameraCtrl \cite{cameractrl} & 84.31 & 93.17 & 97.86 & \textbf{100.00} & 46.26 & 54.78 & 93.27 & 95.05 & 77.52 & 83.09 \\
WorldPlay \cite{hyworld15} & \textbf{95.88} & \underline{94.84} & \textbf{99.29} & 88.00 & 51.76 & 70.92 & \textbf{99.00} & \textbf{99.12} & 83.04 & 87.35 \\
Lingbot \cite{lingbot-world} & 91.82 & 93.93 & 98.01 & \underline{92.00} & \underline{57.09} & \underline{73.30} & 97.08 & 96.64 & \underline{83.66} & \underline{87.48} \\
MoWorld & \underline{95.46} & \textbf{96.33} & 98.12 & \textbf{100.00} & \textbf{59.40} & \textbf{76.65} & \underline{98.22} & \underline{98.33} & \textbf{86.54} & \textbf{90.31} \\
\bottomrule
\end{tabular}}
\end{table}

\begin{figure}[!t]
  \centering
  \includegraphics[alt={},width=\textwidth]{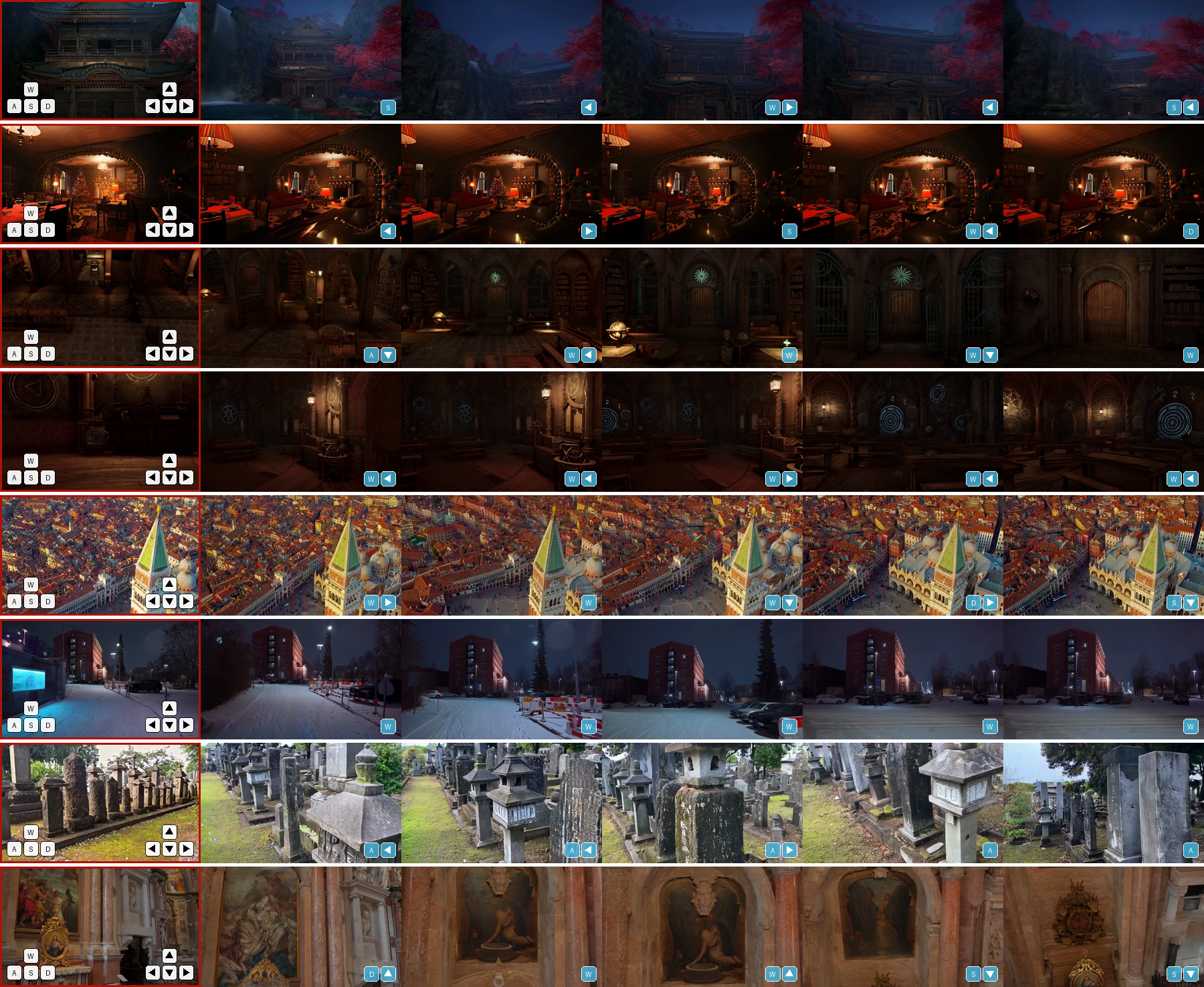}
  \caption{Qualitative results.}
  \label{fig:exp-visualization-2}
\end{figure}

\begin{figure}[!t]
  \centering
  \includegraphics[alt={},width=\textwidth]{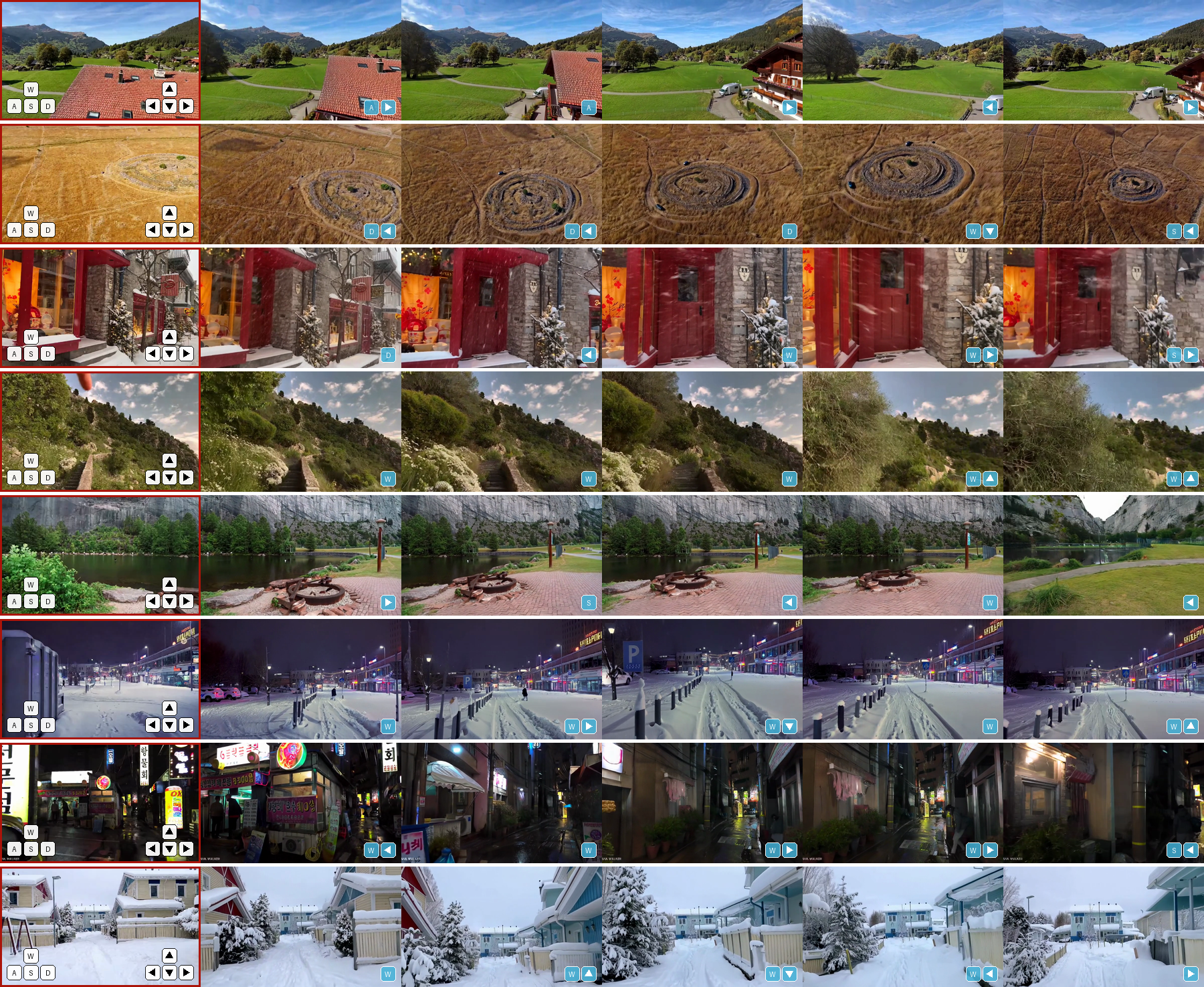}
  \caption{Qualitative results.}
  \label{fig:exp-visualization-2}
\end{figure}

\begin{figure}[!t]
  \centering
  \includegraphics[alt={},width=\textwidth]{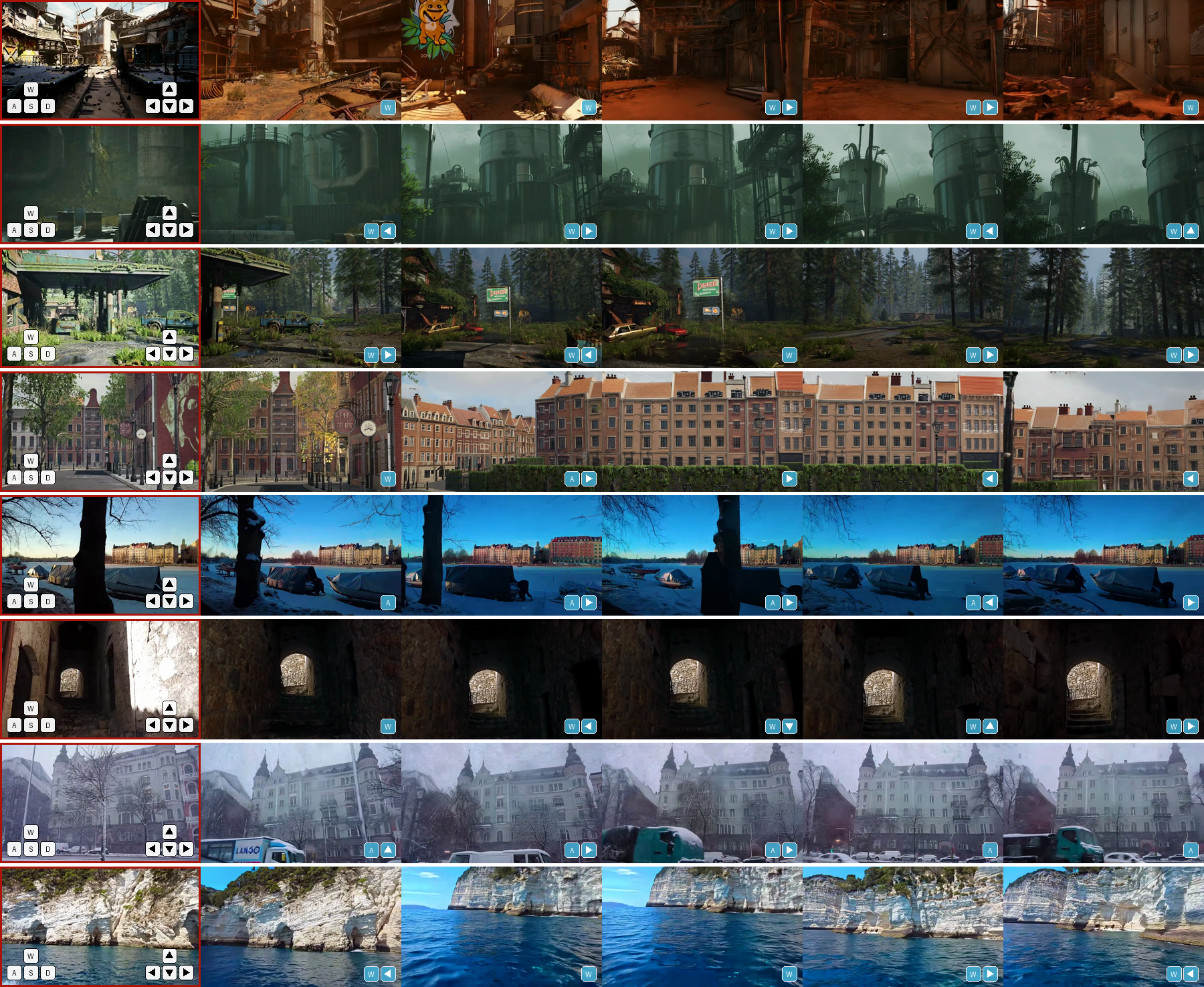}
  \caption{Qualitative results.}
  \label{fig:exp-visualization-2}
\end{figure}

\subsection{Qualitative Results}
\label{sec:qualitative_results}

aTo visually assess the interactive world-simulation capability of MoWorld, we conduct qualitative evaluations across diverse real and virtual scenes. As shown in Fig.~\ref{fig:exp-visualization-1} to Fig.~\ref{fig:exp-visualization-2}, each row corresponds to a different initial environment. The red box denotes the input observation frame, and the subsequent columns show key frames generated under continuous control signals. The overlaid keyboard icons indicate the user action at each time step, including moving forward and backward, translating left and right, and rotating the camera viewpoint.
These visualizations provide an intuitive view of how the model responds to action conditions.

MoWorld can generate continuous and interactive future visual states across diverse scenarios, including indoor libraries, Gothic halls, classical streets, pyramid ruins, shopping-mall corridors, and nighttime residential scenes. The model maintains high visual fidelity while preserving scene structure, material appearance, and lighting style during camera motion. For instance, in indoor scenes, details such as bookshelves, arches, chandeliers, tables, and chairs remain stable after viewpoint changes. In outdoor scenes, large-scale structures such as building facades, rivers, distant skylines, and desert ruins evolve coherently with the camera trajectory.

These results suggest that MoWorld does not merely generate static or weakly dynamic videos. Instead, it continuously updates the visual state according to user actions, forming an interactive world simulation with spatial continuity and action responsiveness. This behavior is particularly evident in geometrically structured environments such as long corridors, indoor rooms, and urban streets, where the model preserves spatial coherence across frames and demonstrates its ability to model the relationship between environment layout and camera motion.

%% file: sections_en/7-application-english.tex
\section{Application}

\begin{figure}[!t]
  \centering
  \includegraphics[alt={},width=0.9\textwidth]{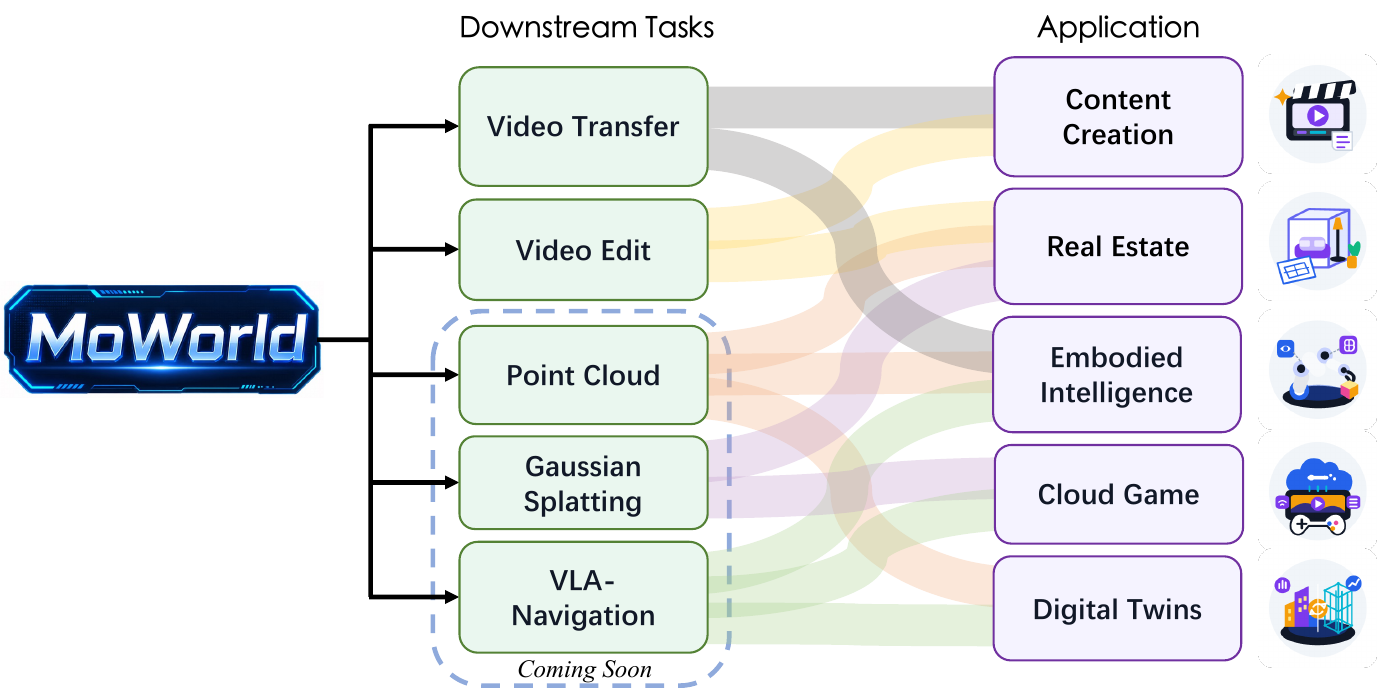}
  \caption{Downstream tasks and applications of MoWorld. MoWorld organizes generated dynamic scenes into five representative downstream tasks, which further support diverse application scenarios such as content creation, real estate visualization, embodied intelligence, cloud gaming, and digital twins.}
  \label{app_graph}
\end{figure}

\begin{figure}[!t]
  \centering
  \includegraphics[alt={},width=\textwidth]{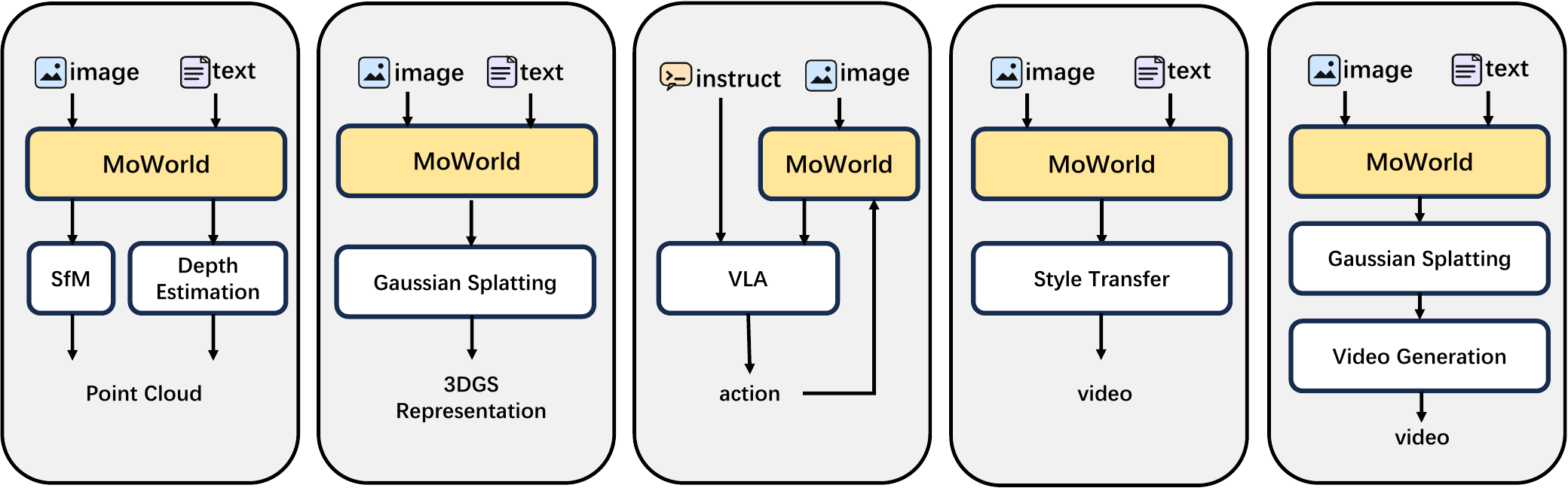}
  \caption{Downstream applications enabled by MoWorld. Given image, text, or instruction inputs, MoWorld can be combined with task-specific modules to support point cloud reconstruction, 3D Gaussian Splatting, navigation, and video generation.}
  \label{fig:app_total}
\end{figure}

\subsection{Overview}
In this work, we introduce MoWorld as a general-purpose world model that learns a unified spatiotemporal representation of dynamic scenes, providing a common foundation for a broad range of downstream tasks. Based on this representation, we organize MoWorld into five representative tasks as shown in Figure~\ref{app_graph}: video transfer, video editing, point cloud reconstruction, 3D gaussian splatting, and navigation.

(1) \textbf{Video Transfer:} In many content creation scenarios, creators often need to reuse the motion dynamics, camera trajectories, or cinematic rhythm of a source video while changing the target scene, style, or visual appearance. MoWorld enables such motion and camera priors to be transferred onto novel scenes or styles, while preserving temporal coherence and geometric consistency throughout the transfer process. This helps creators efficiently generate diverse scene variants without manually redesigning the full motion and camera layout from scratch.

(2) \textbf{Video Editing:} In high-quality film production and visual storytelling, camera motion and narrative progression are often tightly coupled, making it difficult to edit scene content without breaking the overall story flow. By modeling scene evolution together with camera movement, MoWorld supports joint control over both visual content and narrative-consistent camera trajectories, allowing users to insert, remove, or modify objects while maintaining spatially plausible dynamics across frames. This provides creators with a more controllable editing interface for producing coherent cinematic videos with less frame-by-frame manual adjustment.

(3) \textbf{Point Cloud Reconstruction:} For spatial understanding and scene digitization, generated videos alone are often insufficient, since many downstream applications require explicit 3D geometry rather than only RGB observations. The spatially consistent observations produced by MoWorld can serve as geometry-aware multi-view evidence, which can be further converted into point clouds through depth estimation. This helps users obtain explicit 3D scene structures from generated worlds, supporting spatial analysis, real estate visualization, and embodied perception.

(4) \textbf{3D Gaussian Splatting:} Many interactive applications require a real-time renderable 3D representation that can be freely observed from novel viewpoints, rather than a fixed generated video. Since MoWorld produces coherent multi-view observations along controllable camera trajectories, these observations can be lifted into 3D Gaussian Splatting representations for efficient reconstruction and rendering. This helps transform generated dynamic scenes into reusable 3D assets, supporting applications such as cloud gaming, digital twins, and immersive scene visualization.

(5) \textbf{Navigation:} For embodied intelligence, a useful world model should not only generate visual scenes, but also provide an explorable environment for perception, action, and interaction. The generated world can therefore be used as an interactive navigation space, where vision-language-action agents receive observations, follow instructions, execute actions, and interact with the world model in a closed loop. This helps build controllable simulation environments for embodied-agent training, evaluation, and decision-making.

These downstream capabilities naturally extend MoWorld to real-world applications as shown in Figure~\ref{fig:app_total}. By transforming generated dynamic scenes into transferable, editable, reconstructable, and navigable representations, MoWorld can support diverse scenarios such as content creation, real estate visualization, embodied intelligence, cloud gaming, and digital twins. This highlights the potential of MoWorld as a general interface between generative world modeling and spatially grounded interactive applications.

\subsection{Camera-Style Joint Controlling}
Generating diverse visual variants of the same scene while preserving camera motion and spatial layout is essential for video creation and embodied simulation, yet remains costly and labor-intensive.

\begin{figure}
  \centering
  \includegraphics[alt={},width=\textwidth]{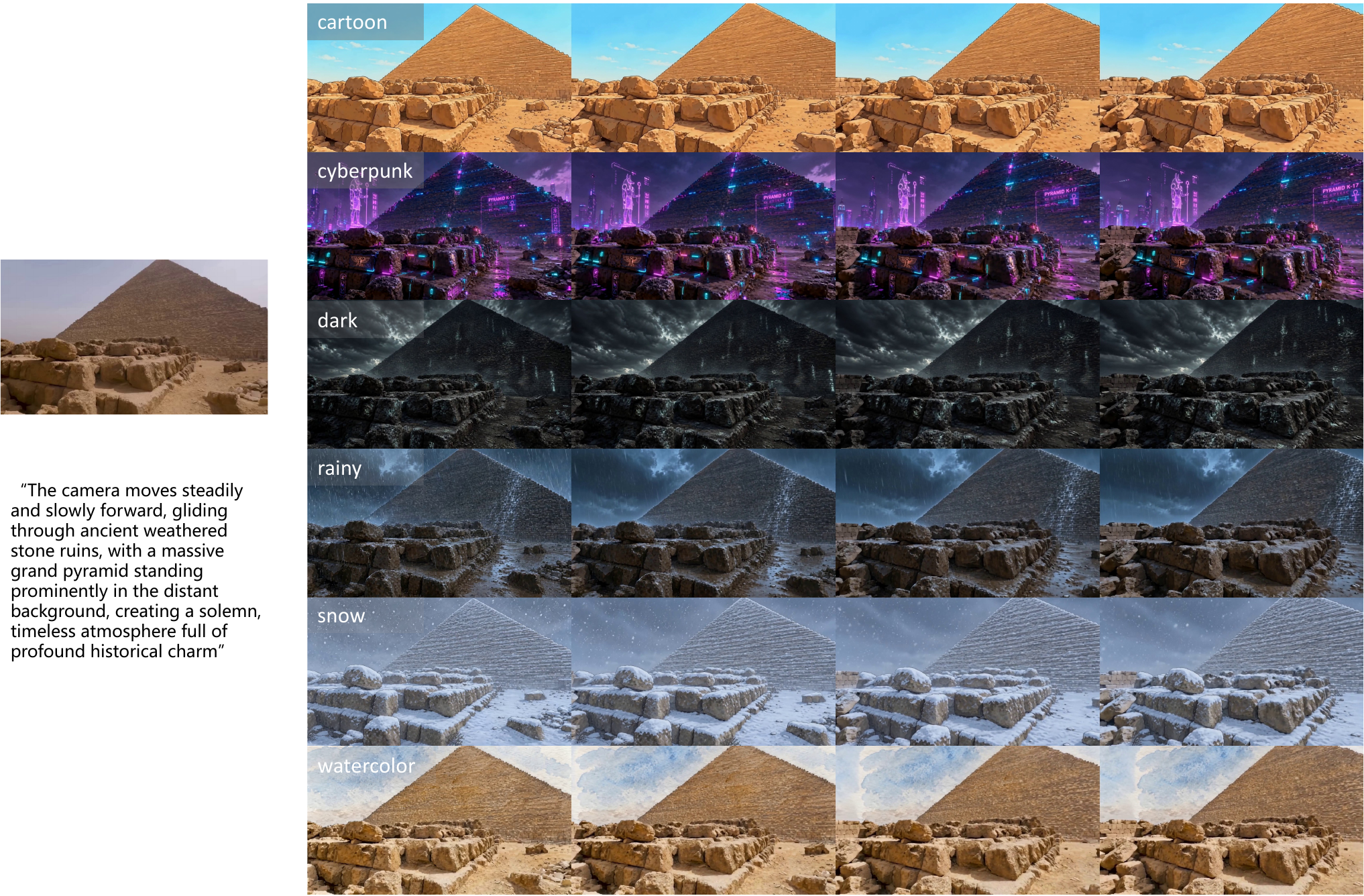}
  \caption{Video transfer results with MoWorld. Given the videos generated by MoWorld, we transfers them into novel styles with consistency motion dynamics and camera trajectory.}
  \label{fig:video-style-transfer}
\end{figure}

By incorporating a video style transfer model~\cite{ali2025world}, MoWorld supports controllable camera trajectories across diverse visual styles. Given an input video, the system can generate stylized variants in different visual domains, including rainy or snowy weather conditions and cyberpunk-style appearances, using text prompts or style-conditioning signals. As shown in Fig.~\ref{fig:video-style-transfer}, MoWorld preserves the original camera motion and spatial layout while transferring the input video into diverse visual styles.

This capability enables several downstream applications. For content creation, it allows creators to efficiently produce multiple stylized versions of the same shot for film story-boarding, game concept design, animation pre-visualization, and short-form video production, thereby reducing the cost of artistic iteration. For embodied intelligence, such stylized videos provide diverse visual training data that exposes agents to variations in weather, illumination, and appearance while preserving the same geometric layout and motion trajectory. This controlled visual diversification enhances agents' perceptual robustness and generalization to complex real-world environments.

\begin{figure}
  \centering
  \includegraphics[alt={},width=\textwidth]{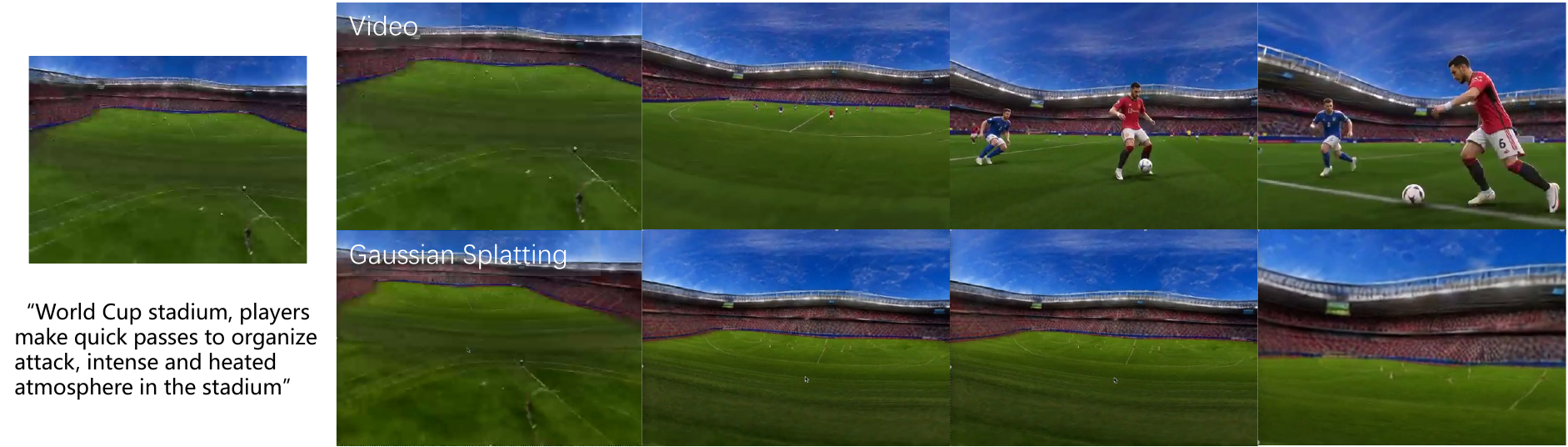}
  \caption{Based on the 3D gaussian splatting results from MoWorld-generated videos, the edited video is obtained with coherent geometry and structure across different camera poses.}
  \label{fig:football}
\end{figure}

\begin{figure}[!t]
  \centering
  \includegraphics[alt={},width=\textwidth]{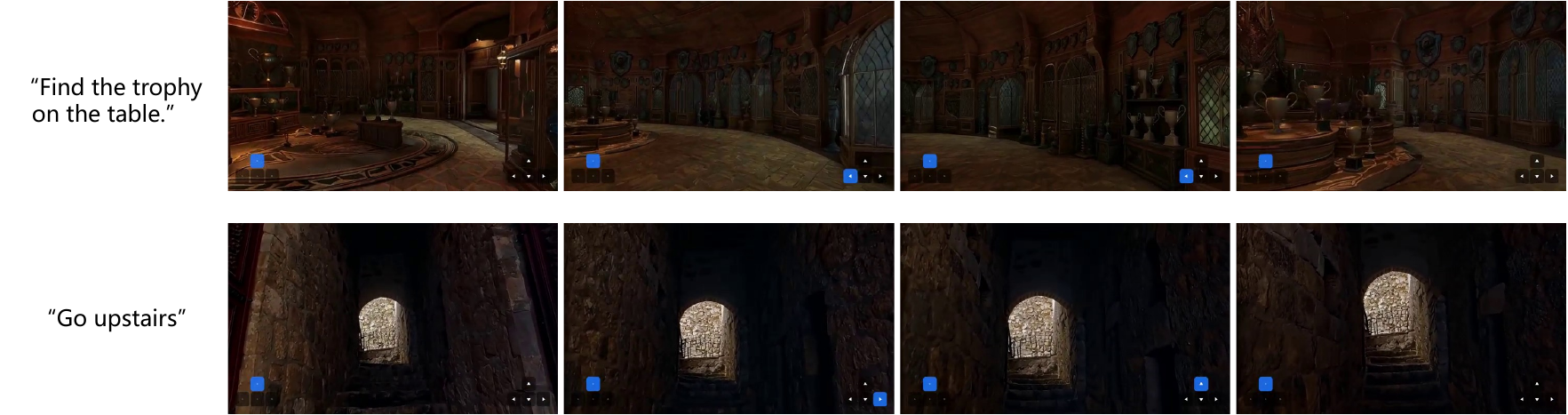}
  \caption{Application to navigation. Given an initial image and a navigation instruction, MoWorld serves as the environment model. The navigation model predicts a sequence of actions, which are then executed in MoWorld to simulate exploration of the scene.}
  \label{fig:app_nav}
\end{figure}

\begin{figure}[!t]
  \centering
  \includegraphics[alt={},width=\textwidth]{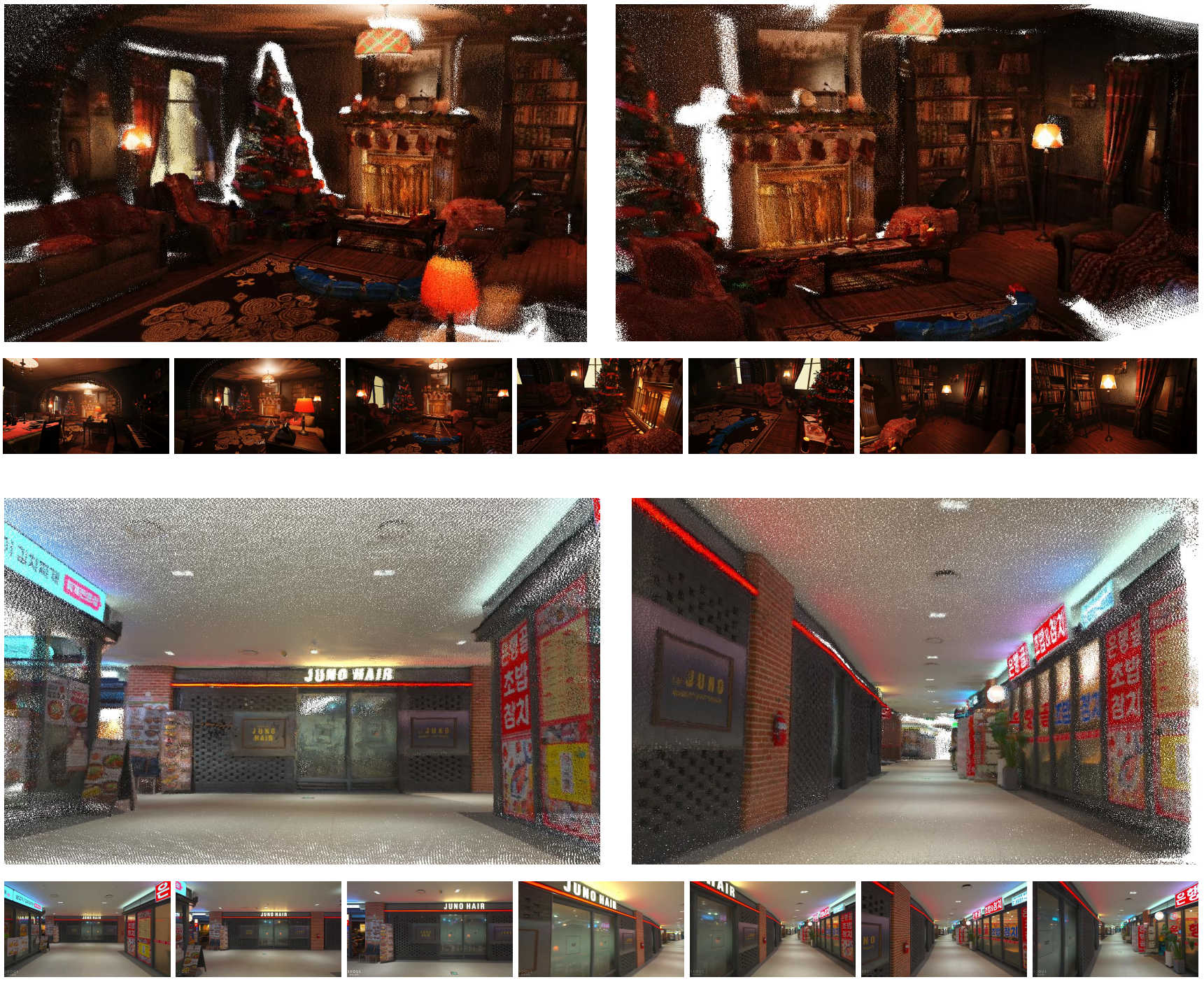}
  \caption{Point cloud reconstruction results from MoWorld-generated videos. Reconstructed point clouds from diverse generated scenes demonstrate coherent spatial structures and reliable geometric consistency across viewpoints.}
  \label{fig:app-pointcloud-1}
\end{figure}

\begin{figure}[!t]
  \centering
  \includegraphics[alt={},width=\textwidth]{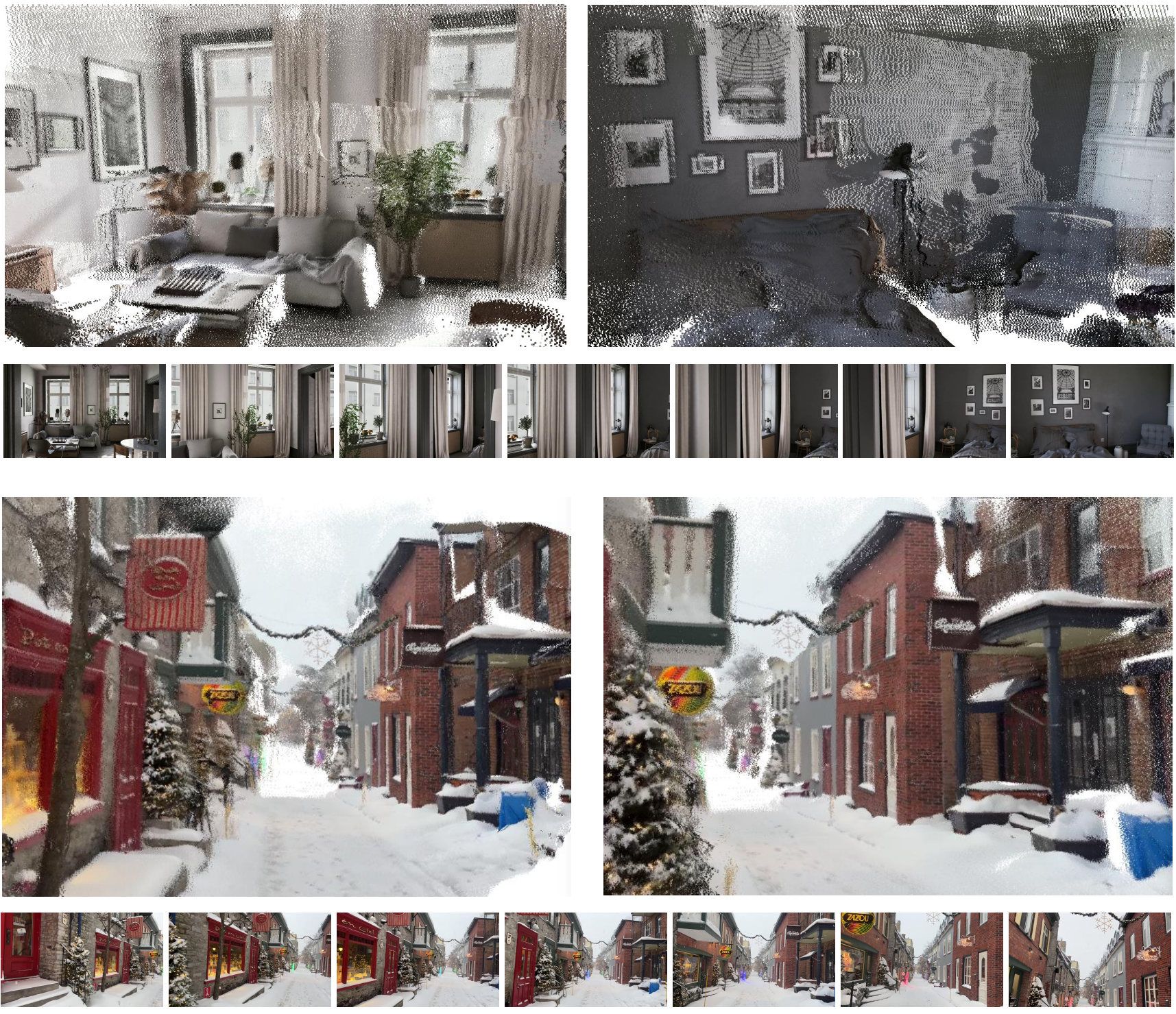}
  \caption{More point cloud reconstruction results from MoWorld-generated videos. The reconstructed point clouds preserve coherent scene layouts and geometric structures across different generated environments, further validating the spatial consistency of MoWorld.}
  \label{fig:app-pointcloud-2}
\end{figure}

\begin{figure}[!t]
  \centering
  \includegraphics[alt={},width=\textwidth]{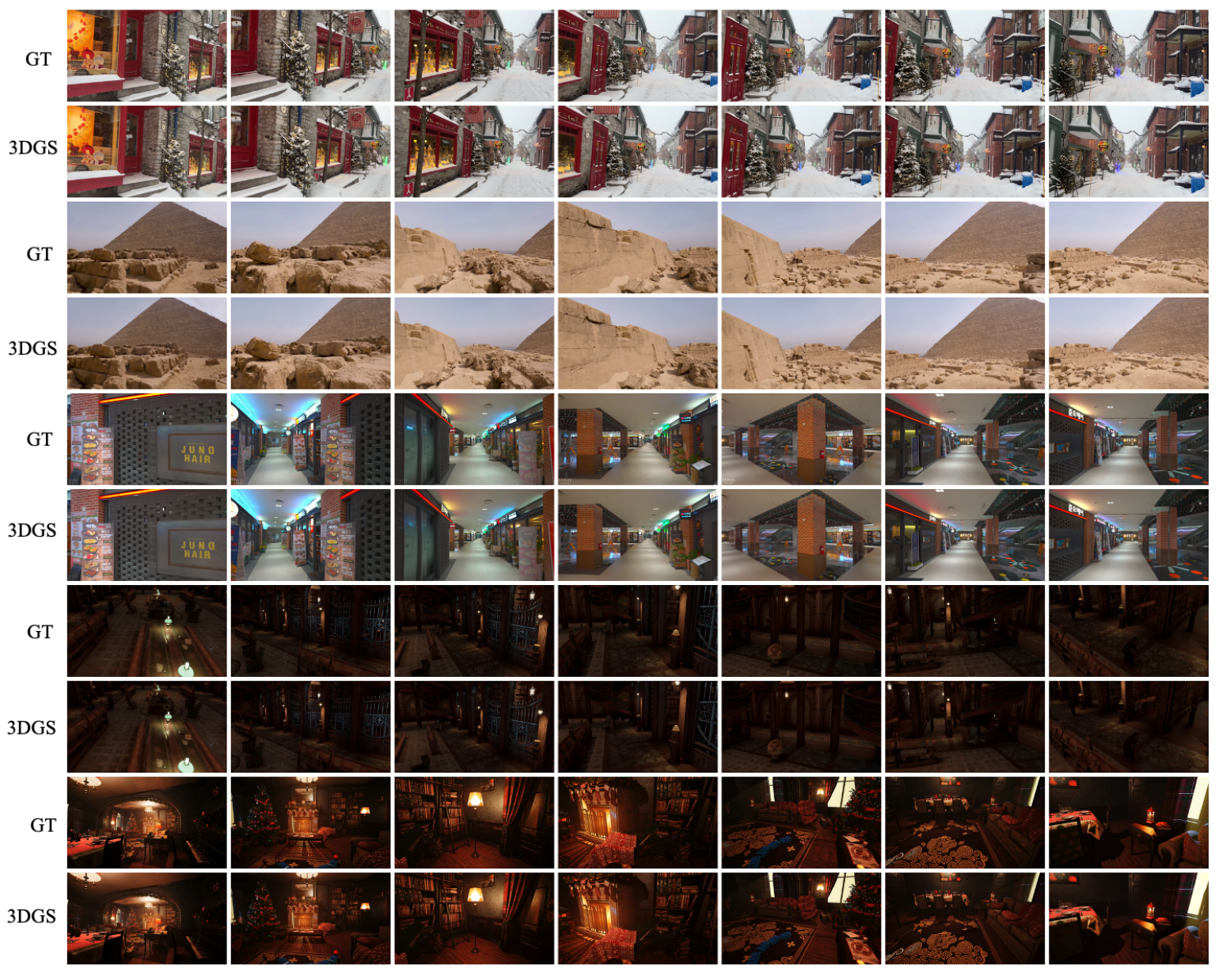}
  \caption{Qualitative comparison of 3D Gaussian Splatting renderings against ground-truth frames. For each scene, the top row shows ground-truth frames extracted from the source video, and the bottom row shows the corresponding novel-view renderings. Results across diverse environments demonstrate strong spatial consistency and geometric fidelity between the reconstructed scenes and the original footage.}
  \label{fig:gaussian}
\end{figure}

\subsection{Camera-Story Joint Controlling}
\label{sec:video-generation-application}

Video generation models provide a new content production paradigm for short dramas, advertisements, and interactive storytelling: users only need to provide scene descriptions and narrative prompts, and the model can generate video clips with camera motion and dynamic events. However, the output duration of existing video generation models is usually limited, and longer stories often need to be decomposed into multiple short clips that are generated separately. Under this setting, relying solely on text prompts makes it difficult to ensure that different clips share the same scene background, which can easily lead to spatial layout drift, discontinuous camera viewpoints, and inconsistent scene identity.
To mitigate this issue, we explicitly introduce scene modeling into the video generation pipeline. The key observation is that ordinary video generation models often lack explicit camera constraints, and the spatial structure and parallax relationships in the generated frames may not satisfy the requirements of 3D reconstruction. In contrast, MoWorld supports controllable camera motion and can generate continuous views of the same scene under a given camera trajectory, thus providing more reliable multi-view observations. Specifically, given a scene video generated by MoWorld, the system first extracts multi-view frames induced by camera-controlled motion and reconstructs the corresponding 3D gaussian splatting scene.
Afterwards, we render a video reference from the reconstructed 3D scene with explicit camera motion and a stable background, and feed it together with text prompts into a video generation model. In this way, the downstream model no longer needs to re-imagine the entire world from text alone. Instead, it mainly focuses on synthesizing dynamic events and foreground motion, such as character actions, interaction processes, and local visual changes. This pipeline effectively decouples ``scene generation'' from ``motion generation'': the reconstructed 3DGS scene provides a stable spatial foundation, while the video model is responsible for generating narrative and dynamic content.

Figure~\ref{fig:football} shows an example of a football scene. The system first reconstructs a 3D representation of the stadium from the football stadium video generated by MoWorld, and obtains a scene video rendered along the camera trajectory. Based on this reference, together with text prompts such as ``World Cup stadium, players make quick passes to organize attack, intense and heated atmosphere in the stadium'', the video model further generates player motion and the atmosphere. As shown in the example, the reconstructed scene provides a stable stadium background for video generation, while the generative model complements the original 3D scene with dynamic characters and events that do not exist in the reconstructed scene.

The advantage of this design is that it turns the structural consistency of large-scale scenes into an explicit constraint, thereby reducing the background drift and viewpoint instability commonly observed in purely text-driven video generation. For scenes with clear spatial layouts, such as stadiums, streets, and indoor spaces, the 3DGS representation reconstructed from scene videos can serve as a reusable 3D environment, enabling the same scene to support video generation under different text prompts, different character motions, and different story versions.

\subsection{Navigation}
\label{sec:navigation-application}

We further demonstrate the use of MoWorld for visual navigation. Given an egocentric observation of the current scene, an action prediction module estimates a short-horizon exploration plan based on the visible environment. The predicted plan is represented as a sequence of coarse control signals, which describe how the agent should move through the scene and adjust its viewpoint during navigation.

These control signals are then converted into a camera trajectory and used to drive MoWorld for video egocentric video sequence generation. In this way, the system~\cite{internvla-n1} first reasons the agent's motion behavior in the control space, and then relies on the world model to synthesize the corresponding visual consequences. This decouples navigation decision making from visual generation: the action module provides the intended movement, while MoWorld renders the resulting egocentric observations under the predicted trajectory.

As shown in Fig.~\ref{fig:app_nav}, the generated egocentric video sequence provide intuitive visual feedback for navigation behavior. The results show that MoWorld can simulate coherent egocentric trajectories under different control signals, making it a useful environment model for exploring navigation-oriented applications.

\subsection{Point Cloud Reconstruction}

Consistent with the benefits of large-scale long-horizon training, MoWorld's generated videos support scene-level point cloud reconstruction. As shown in Figure~\ref{fig:app-pointcloud-1} and Figure~\ref{fig:app-pointcloud-2} , we apply structure-from-motion (SfM)~\cite{vggsfm} and monocular depth estimation~\cite{wang2026vggt} to recover 3D point clouds from the video sequences. The reconstructed point clouds exhibit coherent global structure and consistent spatial layout across the full trajectory, indicating stable cross-frame depth and motion consistency in the generated videos. This provides a geometric foundation for downstream 3D scene understanding and reconstruction tasks.

Such scene-level point clouds unlock a range of practical downstream applications. In real-estate, they can be directly used for spatial measurement and floor-plan generation. In embodied intelligence, the recovered geometry offers a physically grounded scene representation for robot navigation, spatial reasoning, and simulation-to-real transfer. In digital twin construction, point clouds support faithful reconstruction of real-world environments, serving as a scalable data source for facility monitoring and urban scene modeling.

\subsection{Gaussian Reconstruction}

In addition to point cloud reconstruction, MoWorld also supports high-fidelity 3D Gaussian Splatting (3DGS) reconstruction~\cite{kerbl20233d,zhou2024feature} with strong cross-frame geometric coherence. As shown in Figure~\ref{fig:gaussian}, we reconstruct 3DGS representations from the generated video sequences and visualize the rendered frames alongside the corresponding ground-truth frames. The generated videos preserve consistent appearance and structural alignment across frames, allowing the reconstructed 3DGS to closely match the GT in geometry and rendering quality. This demonstrates that MoWorld's outputs carry sufficient cross-frame consistency to support stable 3D reconstruction, helping to alleviate the cross-frame inconsistency commonly observed in conventional video generation models.

The reconstructed 3DGS representations further translate into practical downstream applications. In virtual real estate touring, they allow users to freely navigate and inspect property interiors from arbitrary viewpoints, enabling immersive walkthroughs without costly on-site capture. In cloud gaming, the reconstructed scenes can serve as lightweight, renderable environments that support real-time streaming and interactive exploration, reducing reliance on manually authored 3D assets. These use cases illustrate how MoWorld's generated videos can be converted into render-ready 3D content suitable for real-world deployment.

%% file: sections_en/8-conclusion-english.tex
\section{Conclusion}

MoWorld is a high-performance and cost-efficient world model for interactive video generation and camera-controllable world simulation. From the performance perspective, MoWorld improves controllability, long-horizon consistency, and real-time inference efficiency through scalable data construction, curriculum cross-frame pretraining, efficient distillation, and low-memory, low-latency parallel inference. From a cost perspective, MoWorld reduces both training and deployment costs via curriculum long-video pretraining under limited memory budgets and efficient denoising-step distillation, while enabling real-time prompt interaction across varying NPU compute levels. MoWorld is the first real-time interactive video world model built on the NPU ecosystem. With system-level co-design at the operator, parallelism, and pipeline levels, MoWorld supports pretraining, distillation, and inference for a 14B MoE world model, achieves up to 50 FPS real-time inference, reduces average inference cost to 30\%-50\% of existing world models, and delivers strong results on image-to-video and camera-controllable world-model benchmarks.

%% file: sections_en/9-contribution.tex
\newpage
\section*{Contributors}
\vspace{-0.5mm}

Deyi Ji$^{1\dagger}$*, Tianrun Chen$^{1\ddagger}$*, Xin Zhang$^1$*, Jiale Yang$^1$*, Qi Zhu$^{1\dagger}$*, An Zhao$^2$*, Zihao Xie$^1$*,  Han Wang$^1$*, Xuanyi Liu*, Yixiang Zhou$^2$*, Pei Liu$^1$*, Yi Tan$^1$*, Cheng Chen$^1$*, Dayi Zhu$^1$*,  Mingyu Wei$^1$, Hanjie Xu$^1$, Jun Liao$^1$, Siqi Li$^1$, Lingyu Lu$^1$,  Hongye Fang$^1$, Hongming Tan$^1$, Youjiang Zhu$^1$, Taiyu Zhang$^1$, Zejian Li$^2$, Chaotao Ding$^1$, Zhipeng Liang$^3$, Wenxuan Song$^3$, Yi Li$^3$, Baochuan Yang$^3$, Xin Jiang$^3$, Ben Feng$^3$, Jingyuan Zou$^3$, Yanlin Liu$^3$, Rong Shi$^3$, Lingfeng Li$^3$, Liyi Yao$^3$, Lanyun Zhu$^{\ddagger}$, Yunhe Pan$^2$, Lingyun Sun$^2$.

$^1$\href{http://www.kokoniworld.hk/}{KOKONI 3D, Moxin Technology}, $^2$Zhejiang University, $^3$Huawei

$^\dagger$Project Lead, *Core Technical Contribution, $^\ddagger$Corresponding Authors

\textit{We acknowledge the support from Jianyuan Wang and the support from Huawei Hisilicon Team, Huawei Computing Product Line, and Huawei Cloud Team. Experiments are conducted on Huawei Ascend CloudMatrix384 Supernodes.}